\documentclass[lettersize,journal]{IEEEtran}
\usepackage{booktabs} 
\usepackage{multirow}
\usepackage{stfloats}
\usepackage{amsmath,amsfonts}
\usepackage{algorithmic}
\usepackage{algorithm}
\usepackage{array}
\usepackage{graphicx}
\usepackage[caption=false,font=normalsize,labelfont=sf,textfont=sf]{subfig}
\usepackage{textcomp}
\usepackage{stfloats}
\usepackage{url}
\usepackage{verbatim}
\usepackage{graphicx}
\usepackage{cite}
\begin{document}

\title{Action-Dynamics Modelling and Cross-Temporal Interaction\\ for Online Action Understanding}
\author{Xinyu Yang, Zheheng Jiang, Feixiang Zhou, Yi-hang Zhu, Na Lyu, Nan Xing, Qi Wang and Huiyu Zhou

\thanks{X. Yang, Z. Jiang, Y. Zhu, Q. Wang and H. Zhou are with School of Computing and Mathematical Sciences, University of Leicester, United Kingdom. H. Zhou is the corresponding author. E-mail:hz143@leicester.ac.uk.}
\thanks{F. Zhou is with School of Eye and Vision Sciences, University of Liverpool, United Kingdom.}
\thanks{N. Lyu is with School of Information Science and Engineering, University of Jinan, China.}
\thanks{N. Xing is with School of Automation and Information Engineering, Xi'an University of Technology, China.}
\thanks{This paper has supplementary downloadable material available at http://ieeexplore.ieee.org, provided by the authors. The material includes additional experimental details and results. This material is approximately 0.42 MB in size.}
}

\markboth{Journal of \LaTeX\ Class Files,~Vol.~14, No.~8, August~2021}%
{Shell \MakeLowercase{\textit{et al.}}: A Sample Article Using IEEEtran.cls for IEEE Journals}


\maketitle

\begin{abstract}
Online action understanding, encompassing action detection and anticipation, is central to numerous practical applications. However, untrimmed videos contain substantial redundant and noisy historical information, and the complementary interactions among historical, current, and future-oriented representations remain underexplored. To address these challenges, we propose the State-Specific Model (SSM), a unified framework for online action detection and anticipation. Its Representative State-Based Memory Compression module selects representative anchor frames from the observed memory and aggregates anchor-conditioned context into a compact set of representative memory states, thereby reducing memory redundancy. The subsequent Action Pattern Learning module learns multi-dimensional relations among these states and performs graph reasoning over the resulting directed representative-state graph to model action dynamics and derive a future-oriented cue. The Cross-Temporal Interaction module then enables this cue to interact with historical and current cues, refining the representations used for simultaneous online action detection and anticipation. Extensive experiments on EPIC-Kitchens-100, THUMOS'14, TVSeries, and the introduced Parkinson's Disease Mouse Behaviour dataset demonstrate competitive performance and generalizability across diverse online action understanding settings. These results underscore the value of graph-based action-dynamics modelling and cross-temporal interaction for online action understanding.
\end{abstract}


\section{Introduction} 
\IEEEPARstart{A}{ction} understanding, specifically online action detection \cite{de2016online} and action anticipation \cite{kitani2012activity}, aims to identify current or future actions in streaming videos. In the online setting, only current and historical information is available; future observations are inaccessible. These tasks are fundamental to action retrieval \cite{li2024adaptive}, intelligent surveillance \cite{xia2023exploring}, embodied intelligence (e.g., human–robot interaction \cite{song2018temporal}\cite{li2021restep}), and autonomous driving systems \cite{liu2023hcm}. 
Mainstream approaches predominantly rely on memory mechanisms \cite{xu2021long,  chen2022gatehub, zhao2022real,  girdhar2021anticipative}. For example, the Long Short-term TRansformer \cite{xu2021long} divides its memory encoder into long- and short-term stages for online action detection and anticipation. Other studies extend memory mechanisms in complementary ways. Temporal Smoothing Transformer \cite{zhao2022real} adopts a streaming-transformer paradigm to process long memory sequences and efficiently fuse short- and long-term context for online action detection. Gated History Unit with Background Suppression (GateHub) \cite{chen2022gatehub} applies position-guided gated cross-attention to enhance informative memory segments and suppress background responses. Nevertheless, memory-based models inevitably encounter irrelevant frames during detection and anticipation. This problem becomes more pronounced in longer videos as redundant and noisy information accumulates, potentially obscuring the cues needed to model task-relevant dependencies.



In addition, recent studies on online action understanding have shown that observed and future-oriented cues are complementary and that modelling their relationship can improve performance \cite{gong2022future,xu2019temporal}. However, existing approaches typically exploit this relationship in a predominantly unidirectional manner: they either initialize and update future tokens as placeholders for the final anticipation \cite{gong2022future}, or generate future representations and then use them to support online action detection \cite{xu2019temporal}. Consequently, the representations of historical observations, the current action, and future-oriented context are seldom jointly refined, particularly within a unified detection-and-anticipation framework.

To address these two related challenges, namely, redundant observed memory and insufficient interaction among temporal representations, we propose the State-Specific Model (SSM). SSM follows a compression--relation modelling--interaction pipeline. First, Representative State-Based Memory Compression (RSMC) reduces the redundant observed sequence to a compact set of anchor-conditioned memory states while retaining task-relevant context. Second, Action Pattern Learning (APL) learns multi-dimensional relations among these states and performs graph reasoning to derive a future-oriented cue, i.e., a graph-level latent representation that summarizes observed state dependencies relevant to future-action prediction. Finally, Cross-Temporal Interaction (CTI) allows the historical, current, and graph-derived cues to jointly refine the representations used for online action detection and anticipation. Overall, RSMC primarily addresses memory redundancy, whereas CTI addresses insufficient cross-temporal feature interaction; APL connects the two by transforming the compressed memory states into a relational cue used by CTI. These three constitute complementary stages of the overall SSM framework. Our main contributions are as follows:

\begin{itemize}
\item We propose SSM, a unified framework that jointly addresses redundant observed memory and insufficient cross-temporal representation interaction through a compression--relation modelling--interaction pipeline.
\item We develop a Representative State-Based Memory Compression (RSMC) pipeline in which the ProPos--GMM selection identifies anchors and the proposed Temporal Weighted Attention (TWA) converts these anchors into context-aggregated representative states. The contribution lies in the task-specific compression and state-construction mechanism. 
\item In the proposed Action Pattern Learning (APL) module, we learn multi-dimensional relations among the representative states and perform graph reasoning over the resulting State Transition (ST) Graph. The graph output provides a future-oriented cue that summarizes task-relevant dependencies among the observed states.
\item Our Cross-Temporal Interaction (CTI) module addresses the limited interaction among historical, current, and future-oriented representations through two-stage cross-attention-based feature refinement, enabling complementary online action detection and anticipation within a unified framework.
\item Comprehensive experiments show that SSM achieves competitive performance against representative online action understanding methods across diverse datasets.
\end{itemize}

The remainder of the paper is organized as follows. Section II reviews related work, Section III introduces the proposed method, Section IV reports the experimental results, and Section V concludes the paper.
\vspace{-5pt}
\section{Related Work}
\subsection{Online Action Detection}
Online action detection requires identifying the action occurring at the current time without access to future observations. Because the available information is limited, contemporary approaches commonly rely on memory modelling to exploit historical context. Early methods primarily used recurrent or convolutional architectures to encode the observed history \cite{deo2017learning}. TRN represented past context while also reasoning about possible future information \cite{xu2019temporal}. Discriminative GRU extensions improved the separation of historical action representations \cite{eun2020learning}\cite{cho2014learning}, whereas knowledge distillation was introduced to improve learning efficiency and mitigate the inconsistency of visual content across time \cite{zhao2022progressive}.

Following the success of transformers in temporal sequence modelling \cite{vaswani2017attention}, attention-based architectures have been increasingly adopted for online action understanding. OadTR employed an encoder--decoder framework to jointly encode historical information and detect ongoing actions \cite{wang2021oadtr}. LSTR extended the effective temporal horizon by separately processing long- and short-term memories \cite{xu2021long}. Colar introduced exemplar frames to guide attention towards more discriminative action representations \cite{yang2022colar}, whereas GateHub developed a gated history unit and future-augmented background suppression mechanism to preserve informative temporal cues \cite{chen2022gatehub}. Collectively, these methods demonstrate the importance of long-range historical context. However, their effectiveness still relies on identifying task-relevant evidence from increasingly long observed sequences. As the memory length grows, redundant background content and repeated visual patterns may dilute informative cues, thereby limiting the effectiveness of directly modelling the full temporal sequence.

More recently, MOAD explored efficient extra-long-context modelling through hierarchical scene and action memories, memory quantization, and temporal soft pruning \cite{yan2026backtrace}. Although both MOAD and SSM aim to reduce redundancy in the observed history, they address different aspects of online temporal modelling. MOAD focuses on dynamically maintaining and compressing hierarchical memories, whereas SSM constructs anchor-conditioned representative states and explicitly models the relations among them for both action detection and anticipation. KDSNN, by contrast, targets energy-efficient online video understanding through spiking CNN/Transformer blocks and knowledge distillation \cite{wang2025knowledge}. Thus, KDSNN and SSM focus on different aspects: KDSNN prioritizes energy-efficient deployment, while SSM focuses on temporal abstraction, relational reasoning, and cross-temporal feature interaction. Overall, existing approaches face two persistent challenges: the evidence available at each online inference step is limited, while longer memory sequences introduce increasing amounts of redundancy. SSM addresses the former by facilitating interactions among historical, current, and future-oriented representations, and alleviates the latter by modelling action dynamics rather than relying solely on the full memory sequence.

\subsection{Online Action Anticipation}
Online action anticipation predicts future actions using only the observations available up to the current time. Early work predominantly relied on recurrent networks. The modality-attentive Dual-LSTM encoded and distilled observed sequences to generate rolling predictions, while its learnable attention module fused RGB, optical-flow, and object-centric streams to capture complementary visual cues \cite{furnari2020rolling}. To reduce error accumulation in recurrent prediction, contrastive attention was used to iteratively refine intermediate feature embeddings, with auxiliary verb and noun classification providing additional supervision \cite{qi2021self}. External memory and observed-action classification were subsequently incorporated into the recurrent pipeline, together with contrastive learning that aligned anticipated representations with their ground-truth future counterparts \cite{liu2022hybrid}.

More recent methods have adopted transformer-based architectures to capture longer-range dependencies. AVT combined a transformer encoder over observed video frames with a masked decoder that jointly predicted intermediate and final future representations \cite{girdhar2021anticipative}. Inspired by action recognition, a slow--fast dual-stream design sampled frames at different temporal rates to capture complementary coarse and fine dynamics \cite{osman2021slowfast}. Other work focused on human--object interactions and showed that object-specific cues modelled through attention or transformer modules can indicate which objects are likely to participate in upcoming activities \cite{roy2024interaction}.

Despite these advances, most prior approaches treat action anticipation as an isolated task. They therefore do not fully exploit the complementary temporal information contained in the representations used for online action detection and anticipation: current-action evidence can provide a stronger basis for anticipating what follows, while future-oriented cues inferred from the observed history can help disambiguate the ongoing action. Jointly refining these representations may consequently yield richer and more robust features for both tasks. Building on this insight, SSM performs detection and anticipation simultaneously and uses cross-temporal interaction among historical, current, and future-oriented representations.
\section{Method}
\begin{figure*}[htbp]
  \centering
    \includegraphics[width=1.0\linewidth]{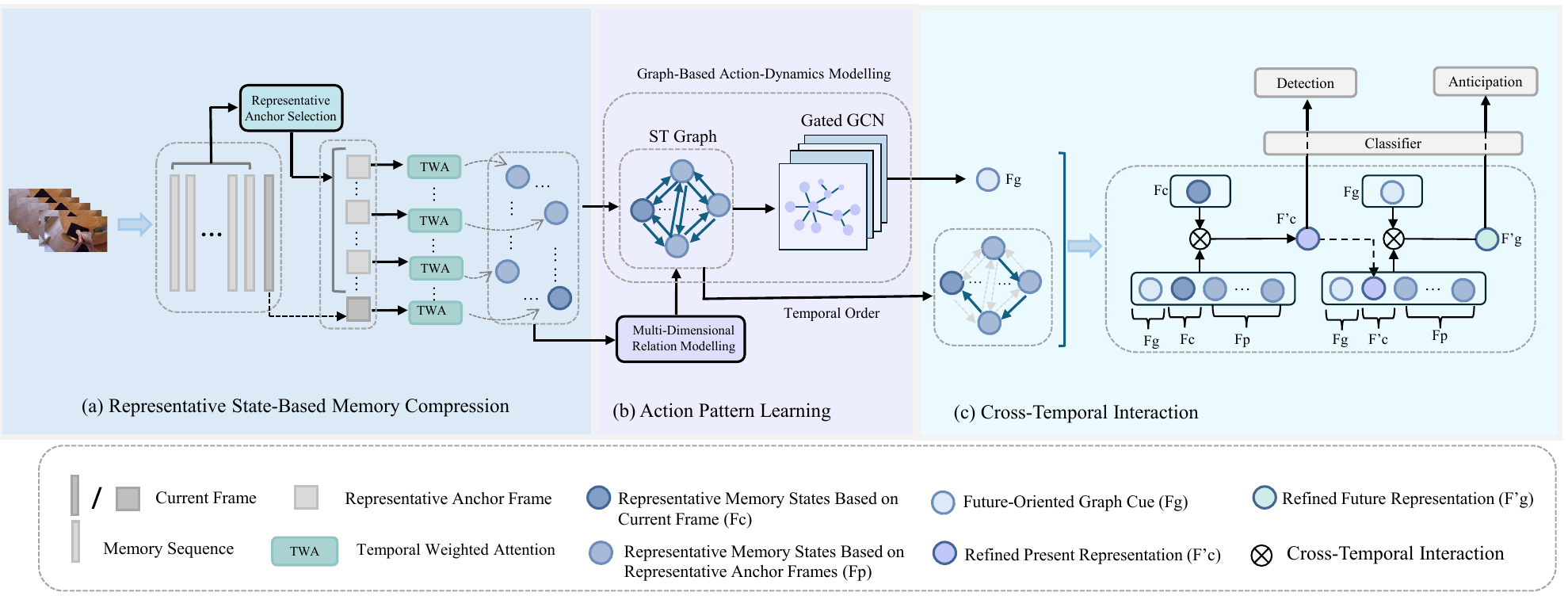}

   \caption{Overview of the proposed State-Specific Model. (a) Representative State-Based Memory Compression. Observed video features are compressed into representative, anchor-conditioned memory states. (b) Action Pattern Learning. The representative states serve as graph nodes, while their learned multi-dimensional relations jointly encode action dynamics. A subsequent Gated Graph Convolutional Network (Gated GCN) propagates and aggregates the relational information to generate a future-oriented graph cue. (c) Cross-Temporal Interaction. Features interact across different time domains to refine current and future cues, supporting action detection and anticipation.}
   \label{fig}
\end{figure*}

As illustrated in Fig.~\ref{fig}, the proposed method performs online action detection and anticipation within a unified streaming framework. At time step $t$, the available modality-specific features are concatenated as $\mathbf{f}_t=[\mathbf{f}_t^{\mathrm{RGB}};\mathbf{f}_t^{\mathrm{Flow}};\mathbf{f}_t^{\mathrm{Obj}}]$, with unavailable modalities omitted according to the experimental setting. For online inference, the model receives only the observed buffer $\mathcal{B}_t=\{\mathbf{f}_{t-L_m},\ldots,\mathbf{f}_{t-1},\mathbf{f}_t\}$, where $\mathbf{F}_m=\{\mathbf{f}_{t-L_m},\ldots,\mathbf{f}_{t-1}\}\in\mathbb{R}^{L_m\times D}$ is the memory sequence, $\mathbf{F}_{\mathrm{current}}=\{\mathbf{f}_t\}$ is the current feature and $D$ denotes the concatenated feature dimension. No feature entry indexed later than $t$ is supplied to the model. The following subsections describe each module and its role in the complete pipeline.
\subsection{Representative State-Based Memory Compression}
To alleviate potential redundancy in the sequential features, we propose the RSMC module. We adopt ProPos \cite{huang2022learning} as a feature-space refinement technique and the Gaussian Mixture Model (GMM) as a window-wise clustering tool for representative-anchor selection. The anchor-selection stage consists of (1) ProPos-GMM clustering of the memory sequence and (2) representative anchor frame selection. In (1), directly applying a generic clustering algorithm to the raw frame features is suboptimal because these features are not necessarily cluster-friendly. We therefore employ ProPos to obtan the instance-consistent embedding with improved intra-cluster compactness and inter-cluster separation. A GMM is then fitted to the ProPos-refined features within each observed buffer. Compared to distance-based hard clustering such as K-means, GMM models both the mean and covariance of each Gaussian component and automatically estimates a mixture coefficient for each cluster. This enables clusters to take elongated, anisotropic shapes that better match the complex geometry of action and background features in untrimmed online videos.
Specifically, given the ProPos-refined feature $\mathbf{f}'_i$ for the $i$-th video frame, the probability density is modelled as $p(\mathbf{f}'_i)=\sum_{k=1}^{K}\pi_k\mathcal{N}(\mathbf{f}'_i\mid\boldsymbol{\mu}_k,\boldsymbol{\Sigma}_k)$, where $K$ is the predefined number of clusters, $\boldsymbol{\mu}_k$ and $\boldsymbol{\Sigma}_k$ denote the mean vector and covariance matrix of the $k$-th Gaussian component, respectively, and $\pi_k$ is the corresponding mixture coefficient automatically estimated through Expectation--Maximization (EM) \cite{dempster1977maximum}. The posterior probability that the $i$-th frame belongs to cluster $k$ is computed by:
\begin{equation}
    p(k\mid\mathbf{f}'_i)=\frac{\pi_k\mathcal{N}(\mathbf{f}'_i\mid\boldsymbol{\mu}_k,\boldsymbol{\Sigma}_k)}{\sum_{j=1}^{K}\pi_j\mathcal{N}(\mathbf{f}'_i\mid\boldsymbol{\mu}_j,\boldsymbol{\Sigma}_j)}
  \label{2}
\end{equation}

After clustering, step (2) maps each Gaussian component to an actually
observed frame. Each observed feature is first assigned to the component with
the largest posterior probability. Let
\begin{equation}
\mathcal{I}_k=
\left\{i \,\middle|\,
k=\arg\max_{l\in\{1,\ldots,K\}}p(l\mid\mathbf{f}'_i)\right\}
\label{eq:component_membership}
\end{equation}
denote the indices assigned to the $k$-th component. Its representative
anchor index and corresponding input feature are selected by
\begin{equation}
r_k=\arg\min_{i\in\mathcal{I}_k}
\left\|\mathbf{f}'_i-\boldsymbol{\mu}_k\right\|_2,
\qquad
\mathbf{x}_k^{\mathrm{anc}}=\mathbf{f}_{r_k}.
\label{eq:euclidean_anchor}
\end{equation}
That is, the frame closest to the mean is selected as the anchor for each cluster. Finally, the $K$
historical anchors are combined with the current frame
$\mathbf{x}_{K+1}^{\mathrm{anc}}=\mathbf{f}_t$ to form
$\mathcal{C}=\{\mathbf{x}_1^{\mathrm{anc}},\ldots,
\mathbf{x}_K^{\mathrm{anc}},\mathbf{x}_{K+1}^{\mathrm{anc}}\}$.

The selected anchor frames provide representative samples of the observed
feature distribution, but relying on the anchors alone may overlook useful
contextual information. To address this limitation, we propose Temporal
Weighted Attention (TWA), in which the representative anchor frames serve as
queries $(\mathbf{Q})$ and the observed memory-current sequence ($\mathcal{B}_t$) serves as keys $(\mathbf{K})$
and values $(\mathbf{V})$. Let
$\Delta t_{i,j}=\lvert t_i-t_j\rvert$ denote the temporal distance between
the $i$-th anchor and the $j$-th observed frame. We use the Gaussian temporal
kernel
$g_{\mathrm{G}}(\Delta t_{i,j})
=\exp(-\Delta t_{i,j}^{2}/(2\delta^{2}))$, where $\delta$ controls the
effective temporal neighbourhood around an anchor. 

Let $s_{i,j}=\mathbf{Q}_i\mathbf{K}_j^{\top}/\sqrt{d_k}$ denote the
content-based compatibility score, where $d_k$ is the query/key dimension.
TWA modulates this score using the temporal kernel and applies a softmax over
the observed sequence:

\begin{equation}
\begin{split}
a_{i,j}
&=
\frac{\exp(s_{i,j})g_{\mathrm{G}}(\Delta t_{i,j})}
{\sum_{l=0}^{L_m}
\exp(s_{i,l})g_{\mathrm{G}}(\Delta t_{i,l})},\\
\mathbf{S}_i
&=\sum_{j=0}^{L_m}a_{i,j}\mathbf{V}_j.
\end{split}
\label{eq:attention}
\end{equation}
The Gaussian form provides a smooth, symmetric, and monotonically decreasing locality prior without introducing a hard temporal boundary. Its squared-distance decay suppresses potentially irrelevant distant observations in the long memory buffer, while the content-based attention scores continue to determine which observations contribute to the state. The resulting content--time weighting emphasizes local evidence around each anchor but can retain distant frames when supported by their content compatibility. Ultimately, TWA compresses the sequence into $K+1$
representative memory states. Each state is an anchor-conditioned,
context-aggregated feature constructed from the observed sequence and serves
as a compact node representation for the subsequent APL module. An individual
state is not intended to encode action progression, an action boundary, or a
future event; dependencies among states are modelled by the relations
and message propagation in APL.
\subsection{Action Pattern Learning}
The representative states obtained by RSMC are anchored at different observed time points and summarize their corresponding temporal contexts. Although an individual state is not assumed to represent a predefined action phase, the dependencies among these temporally grounded states characterize how the observed action cues relate and evolve over time. In this work, we operationally define this evolving, task-relevant dependency structure among representative states as action dynamics. It is represented jointly by temporally grounded node states, directed multi-dimensional inter-state relations, and edge-gated graph propagation, rather than by assigning an explicit semantic meaning to an individual state or relation. The formulation and construction of these components are detailed in the remainder of this section.

To represent such dynamics, APL constructs directed multi-dimensional relations between every pair of representative states. Given two states $\mathbf{S}_i$ and $\mathbf{S}_j$, APL estimates $\mathbf{E}_{i,j}$ and $\mathbf{E}_{j,i}$ separately. Unlike a weighted, fully connected graph in which each edge is represented by a single scalar expressing one predefined relation, such as temporal adjacency, our multi-dimensional edges are learnable vectors capable of encoding complementary task-relevant dependencies between the source and target states, as suggested in \cite{luo2022learning}. The channel-wise semantics are not manually specified; instead, the edge representations are conditioned on the complete set of representative states and are learned jointly under the downstream online action understanding objectives. 

Let $\mathbf{S}=[\mathbf{S}_1;\ldots;\mathbf{S}_{K+1}]\in\mathbb{R}^{(K+1)\times d}$ denote the representative states. For the $h$-th attention head, where $h\in\{1,\ldots,H\}$, the query, key, and value associated with an ordered state pair $(i,j)$ are
\begin{equation}
\mathbf{q}_{i}^{h}=\mathbf{W}_{q}^{h}\mathbf{S}_{i},\qquad
\mathbf{k}_{j}^{h}=\mathbf{W}_{k}^{h}\mathbf{S}_{j},\qquad
\mathbf{v}_{j}^{h}=\mathbf{W}_{v}^{h}\mathbf{S}_{j}.
\label{eq:edge_qkv}
\end{equation}
The pair-specific attention coefficient and corresponding head-wise relation feature are computed as
\begin{equation}
\alpha_{i,j}^{h}
=
\frac{\exp\bigl((\mathbf{q}_{i}^{h})^{\top}\mathbf{k}_{j}^{h}/\sqrt{d_h}\bigr)}
{\sum_{l=1}^{K+1}\exp\bigl((\mathbf{q}_{i}^{h})^{\top}\mathbf{k}_{l}^{h}/\sqrt{d_h}\bigr)},
\qquad
\mathbf{z}_{i,j}^{h}=\alpha_{i,j}^{h}\mathbf{v}_{j}^{h},
\label{eq:edge_attention}
\end{equation}
where $d_h$ is the feature dimension of each head. The outputs of the $H$ heads are concatenated and linearly projected to obtain the directed multi-dimensional edge
\begin{equation}
\mathbf{E}_{i,j}
=
\mathbf{W}_{e}
[\mathbf{z}_{i,j}^{1};\ldots;\mathbf{z}_{i,j}^{H}]
\in\mathbb{R}^{d_e}.
\label{eq:edge_encoding}
\end{equation}
where $d_e$ is the edge-feature dimension. Because the source state determines the query and the attention is normalized over its candidate target states, generally $\mathbf{E}_{i,j}\ne\mathbf{E}_{j,i}$; hence, the two directions are estimated separately.

The representative states and their directed multi-dimensional relations constitute the ST Graph, a learned relational graph over temporally grounded states rather than a semantically annotated state-transition system. The adjacency matrix $\mathbf{A} \in \{0,1\}^{(K+1) \times (K+1)}$ is defined as follows: $A_{i,j}=1$ if $i\ne j$, and $A_{i,j}=0$ otherwise. Here, $A_{i,j}=1$ indicates a directed edge from node $i$ to node $j$. 


Once constructed, the ST Graph is processed by a Gated Graph Convolutional Network (Gated GCN) \cite{bresson2017residual}. We initialize $\mathbf{h}_{i}^{(0)}=\mathbf{S}_{i}$ and $\mathbf{e}_{i,j}^{(0)}=\mathbf{E}_{i,j}$. At layer $l$, the edge-conditioned gate and the message aggregated at target node $j$ are
\begin{equation}
\mathbf{g}_{i,j}^{(l)}
=\sigma\bigl(\mathbf{W}_{\mathrm{gate}}^{(l)}\mathbf{e}_{i,j}^{(l)}\bigr),
\qquad
\mathbf{m}_{j}^{(l)}
=\sum_{i:A_{i,j}=1}
\mathbf{g}_{i,j}^{(l)}
\odot
\bigl(\mathbf{W}_{h}^{(l)}\mathbf{h}_{i}^{(l)}\bigr),
\label{eq:gated_message}
\end{equation}
where $\sigma(\cdot)$ is a sigmoid function and $\odot$ denotes element-wise multiplication. Thus, each multi-dimensional edge regulates the channel-wise message transmitted from its source state $i$ to its target state $j$. Let $\{\mathbf{h}_{i}^{(L_g)}\}_{i=1}^{K+1}$ denote the node embeddings after $L_g$ Gated GCN layers. We obtain the graph-level representation by global average pooling:
\begin{equation}
\mathbf{F}_g=\frac{1}{K+1}\sum_{i=1}^{K+1}\mathbf{h}_{i}^{(L_g)}.
\label{eq:graph_cue}
\end{equation}
An auxiliary future-action distribution is then produced from this representation:
\begin{equation}
\mathbf{p}_g=\operatorname{Softmax}(\mathbf{W}_g\mathbf{F}_g+\mathbf{b}_g),
\label{eq:graph_prediction}
\end{equation}
where $\mathbf{W}_g$ and $\mathbf{b}_g$ are learnable parameters. Because $\mathbf{F}_g$ is derived solely from the action dynamics and is optimized to contain information relevant to future actions through the anticipation and future-distribution consistency objectives, we refer to it as a \emph{future-oriented graph cue}. Note this term describes its operational role and does not imply that $\mathbf{F}_g$ represents an explicitly annotated or psychologically grounded human intention. 


\subsection{Cross-Temporal Interaction}
\begin{figure}[htbp]
  \centering
    \includegraphics[width=0.7\linewidth]{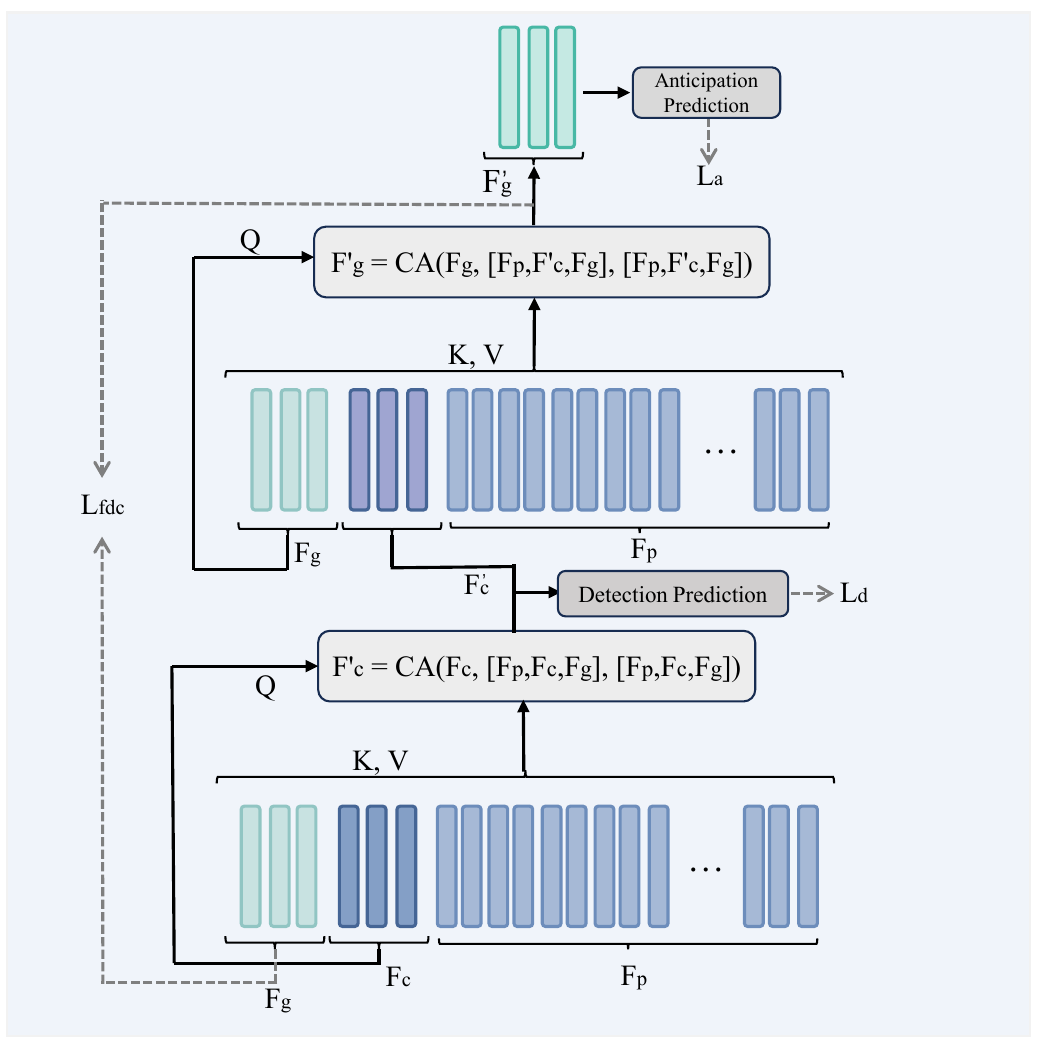}
   \caption{CTI models the interaction between the future-oriented graph cue and past-present cues. This process further refines the temporal representations, supporting online action detection and anticipation.}
   \label{fig3}
\end{figure}
The future-oriented graph cue represents broad and abstract action trends. Let the output of RSMC be $\mathbf{S}=[\mathbf{S}_1;\ldots;\mathbf{S}_K;\mathbf{S}_{K+1}]\in\mathbb{R}^{(K+1)\times d}$, where $\mathbf{S}_1,\ldots,\mathbf{S}_K$ are the representative states constructed from the historical anchors and $\mathbf{S}_{K+1}$ is the representative state constructed using the current frame $\mathbf{f}_t$ as its anchor. The three inputs to CTI are defined as
\begin{equation}
\begin{aligned}
\mathbf{F}_p&=[\mathbf{S}_1;\ldots;\mathbf{S}_K],\\
\mathbf{F}_c&=\mathbf{S}_{K+1},\\
\mathbf{F}_g&=\frac{1}{K+1}\sum_{i=1}^{K+1}\mathbf{h}_i^{(L_g)} ,
\end{aligned}
\label{eq:cti_inputs}
\end{equation}
where $\mathbf{F}_p\in\mathbb{R}^{K\times d}$ and $\mathbf{F}_c,\mathbf{F}_g\in\mathbb{R}^{d}$ denote the past cue, present cue, and future-oriented graph cue, respectively. All three cues are computed exclusively from observations available no later than time $t$. 

We employ cross-attention $CA(Q,K,V)$ to refine the cross-temporal representations. The two CTI updates are formulated as
\begin{equation}
\mathbf{F}_t=[\mathbf{F}_p;\mathbf{F}_c;\mathbf{F}_g],\qquad
\mathbf{F}'_c=CA(\mathbf{F}_c,\mathbf{F}_t,\mathbf{F}_t),
\label{eq:cti_current}
\end{equation}
\begin{equation}
\mathbf{F}'_t=[\mathbf{F}_p;\mathbf{F}'_c;\mathbf{F}_g],\qquad
\mathbf{F}'_g=CA(\mathbf{F}_g,\mathbf{F}'_t,\mathbf{F}'_t).
\label{eq:cti_future}
\end{equation}
These two updates are executed once in a feed-forward order: the present representation is refined first, after which the refined present representation is used to update the future-oriented representation. Thus, CTI is a two-stage cross-attention-based feature-refinement mechanism rather than an iterative feedback loop, recurrent reasoning process, or causal closed-loop model, as shown in Fig. ~\ref{fig3}.

The final prediction head is a shared linear classifier. The current- and future-action distributions are computed as
\begin{equation}
\begin{aligned}
\mathbf{p}_d&=\operatorname{Softmax}
(\mathbf{W}_{\mathrm{cls}}\mathbf{F}'_c+\mathbf{b}_{\mathrm{cls}}),\\
\mathbf{p}_a&=\operatorname{Softmax}
(\mathbf{W}_{\mathrm{cls}}\mathbf{F}'_g+\mathbf{b}_{\mathrm{cls}}),
\end{aligned}
\label{eq:task_predictions}
\end{equation}
where $\mathbf{W}_{\mathrm{cls}}$ and $\mathbf{b}_{\mathrm{cls}}$ are shared between detection and anticipation. $\mathbf{F}'_c$ and $\mathbf{F}'_g$ are used for current-action detection and future-action anticipation, respectively. 

\subsection{Loss Function}
To jointly optimize online action detection and anticipation while encouraging agreement between the graph-derived and CTI-refined future predictions, we design a multi-component loss function:

\textbf{Action Detection Loss $L_{d}$}: To accurately identify ongoing actions within the current frame, we employ a supervised cross-entropy (CE) loss defined as $L_d=CE(\mathbf{y}_d,\mathbf{p}_d)$, where $\mathbf{y}_d$ denotes the current-action label and $\mathbf{p}_d$ represents the predicted probability distribution for current actions.

\textbf{Action Anticipation Loss $L_{a}$}: To facilitate precise anticipation of future actions, we define the cross-entropy loss $L_a=CE(\mathbf{y}_a,\mathbf{p}_a)$, where $\mathbf{y}_a$ denotes the future-action label at the evaluated anticipation horizon and $\mathbf{p}_a$ denotes the corresponding predicted distribution.

\textbf{Future-Distribution Consistency Loss $L_{\mathrm{fdc}}$}: Let $c\in\{1,\ldots,C\}$ denote a future-action category. The auxiliary probability $p_g(c)$ is produced directly from the graph cue $\mathbf{F}_g$ as defined in Eq.~\eqref{eq:graph_prediction}, while $p_a(c)$ is produced from the CTI-refined future representation $\mathbf{F}'_g$ as defined in Eq.~\eqref{eq:task_predictions}. Both distributions predict the same future-action label space at the same anticipation horizon. We encourage agreement between these two future predictions using
\begin{equation}
L_{\mathrm{fdc}}=D_{\mathrm{KL}}(\mathbf{p}_g\parallel\mathbf{p}_a)
=\sum_{c=1}^{C}p_g(c)\log\frac{p_g(c)}{p_a(c)}.
\label{eq:future_consistency}
\end{equation}
This term is a distribution-level consistency regularizer; it does not encode explicit logical rules, symbolic constraints, or psychologically grounded human intention.

Overall, our complete optimization objective is a weighted combination of these three terms: $L=L_d+\lambda_1L_a+\lambda_2L_{\mathrm{fdc}}$, where $\lambda_1$ and $\lambda_2$ balance current-action detection, future-action anticipation, and distribution-level agreement between the graph-derived and CTI-refined future predictions. By jointly optimizing these terms, our method learns representations that support both accurate online action detection and anticipation.
\section{Experiments}
\subsection{Datasets, Metrics and Implementation Details}
\textbf{Datasets.} We evaluate the proposed method on three public benchmark datasets: EPIC-Kitchens-100 \cite{damen2022rescaling}, THUMOS'14 \cite{THUMOS14}, and TVSeries \cite{de2016online}. Further details on these datasets are provided in Section I (Datasets) of the supplementary material.

To evaluate generalization in a specialized domain, we introduce the Parkinson's Disease Mouse Behaviour (PDMB) dataset \cite{zhou2024smc}. Previous studies \cite{zhou2025cross, yang2024online} have demonstrated its reliability for investigating mouse behavioural patterns associated with Parkinson's disease. The dataset contains videos of approximately 9--14 minutes, with frame-level annotations for eight action categories. Following prior work \cite{zhou2025cross, yang2024online}, we consider eight social behaviours: approach, chase, circle, eat, clean, sniff, up, and walk away. We use eight videos for training and two for testing; both splits contain instances of every behaviour category, supporting evaluation in a medically relevant behavioural-analysis setting.

\textbf{Metrics.} For THUMOS'14, we evaluate performance using mean Average Precision (mAP). For TVSeries, we adopt mean calibrated Average Precision (mcAP) \cite{de2016online}. For EPIC-Kitchens-100, we follow the evaluation protocol established in \cite{damen2022rescaling} and report class-mean top-5 recall separately for verbs, nouns, and actions. For PDMB, we use both mAP and mcAP. For action anticipation on EPIC-Kitchens-100, we follow the standard evaluation protocol \cite{damen2022rescaling, guo2024uncertainty, wang2023memory} and primarily evaluate an anticipation time gap of $t = 1$ s; the same anticipation gap is adopted for PDMB. For THUMOS'14 and TVSeries, we evaluate anticipation at multiple time gaps from 0.25s to 2.0s in steps of 0.25s.

\textbf{Implementation Details.}
For EPIC-Kitchens-100, we use RGB, optical flow, and object features; for TVSeries, THUMOS'14, and PDMB, we use only RGB and optical flow features. Following the standard feature-based online protocol of prior work \cite{xu2021long,zhao2022real,wang2023memory}, every temporal-model prediction at step $t$ uses only pre-extracted feature entries indexed no later than $t$. RSMC, APL and CTI introduce no additional look-ahead. The raw-frame temporal support of each input follows the corresponding feature-extraction protocol; details for every modality are provided in Section II (Feature Extraction and Online Protocol) of the supplementary material. We train the model using Adam with an initial learning rate of $3\times 10^{-4}$ and a weight decay of $1\times 10^{-4}$. For the temporal weighting function in RSMC, we set $\delta=1$. The loss-balancing hyperparameters $\lambda_{1}$ and $\lambda_{2}$ are set to 1.1 and 0.7, respectively, controlling the contributions of detection, anticipation, and future-distribution consistency. Further details of the ProPos and GMM configuration are provided in Section VIII of the supplementary material. All experiments were conducted on the Sulis High-Performance Computing (HPC) platform.
\subsection{Ablation Study}
\begin{figure}[htbp]
  \centering
    \includegraphics[width=1.0\linewidth]{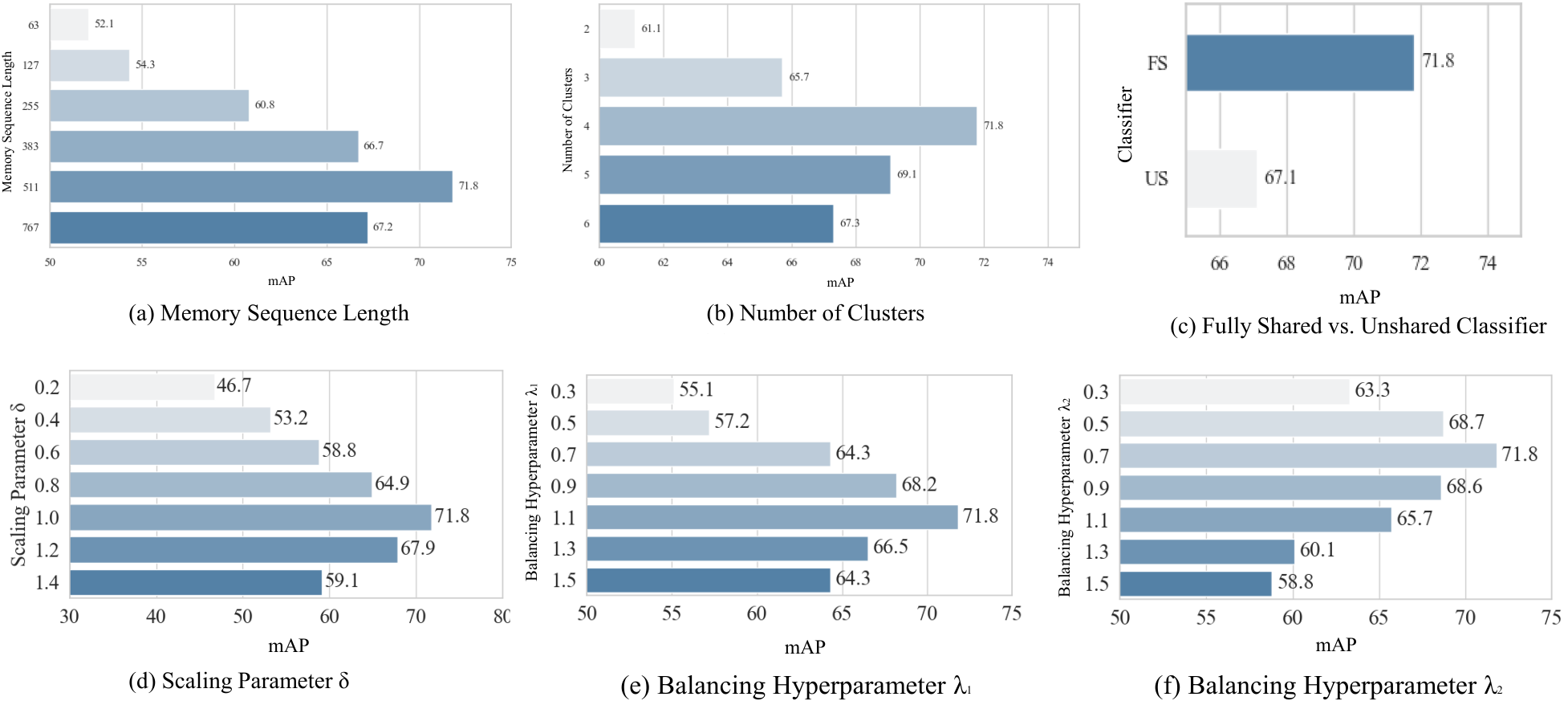}

   \caption{Ablation study on (a) memory sequence length; (b) number of clusters; (c) fully shared (FS) classifier vs. unshared (US) classifier; (d) scaling parameter $\delta$; (e) balancing hyperparameter $\lambda_1$; and (f) balancing hyperparameter $\lambda_2$. All these experiments are on the THUMOS'14 test set and mAP is adopted as the evaluation metric.}
   \label{fig:ablation_overview}
\end{figure}

\textbf{Memory Sequence Length.} Fig. 3(a) studies the impact of the memory sequence length (the buffer excluding the current frame) $L_m$ in RSMC. Increasing $L_m$ improves performance at first by providing richer temporal context and action cues, but the gain reverses when $L_m$ exceeds $511$. Longer sequences introduce irrelevant frames, which add noise to the compressed representations, while also increasing computation and making it harder to model relations among representative states. Thus, we set $L_m=511$.

\textbf{Number of Clusters.} As shown in Fig. 3(b), performance improves as the number of clusters ($K$) increases, but drops after $K=4$. A moderate $K$ provides a compact yet sufficiently diverse set of representative nodes for subsequent relation modelling, whereas a larger $K$ makes the graph unnecessarily complex: dependencies are diluted by dense connections, and node contributions become less distinctive, causing ambiguity and lower accuracy. Therefore, we set $K=4$.

\textbf{Shared Classifier.} Fig. 3(c) compares classifier-sharing strategies for action detection and anticipation. A fully shared classifier achieves the best overall performance, as it enforces a unified decision boundary for consistent predictions across tasks and aligns complementary cross-temporal features into a shared discriminative semantic space, enhancing task complementarity.

\textbf{Parameter Selection.}
To identify optimal hyperparameters, we conduct ablations on three key parameters (Fig. 3(d–f)). The scaling factor $\delta$ in Temporal Weighted Attention controls the temporal weighting window. As shown in Fig. 3(d), moderate $\delta$ improves performance, while too small $\delta$ over-emphasizes local frames and too large $\delta$ over-smooths temporal cues; thus, we set $\delta=1$ for the best locality–context trade-off. We further tune the balancing hyperparameters $\lambda_1$ and $\lambda_2$ (Fig. 3(e–f)). Very small $\lambda$ weakens the corresponding terms, whereas overly large $\lambda$ disrupts optimization balance and degrades performance. We therefore choose $\lambda_1=1.1$ and $\lambda_2=0.7$.

\begin{table}[t]
\centering
\caption{Ablation of temporal weighting kernels on the THUMOS'14 test set
(mAP, \%). }
\label{tab:temporal_kernel}
\footnotesize
\setlength{\tabcolsep}{5pt}
\begin{tabular}{@{}lcc@{}}
\toprule
Temporal weighting & Detection & Anticipation \\
\midrule
None (vanilla attention) & 69.8 & 59.3 \\
Gaussian                 & \textbf{71.8} & \textbf{62.6} \\
Exponential              & 70.9 & 61.3 \\
Cauchy                   & 71.1 & 60.7 \\
\bottomrule
\end{tabular}
\end{table}

\textbf{Temporal Weighting Kernel.}
Table~\ref{tab:temporal_kernel} evaluates whether the benefit of TWA depends
on the Gaussian temporal prior. In addition to vanilla attention without a
temporal prior, we compare the Gaussian kernel with the exponential kernel
$g_{\mathrm{E}}(d)=\exp(-d/\delta)$ and the heavier-tailed Cauchy kernel
$g_{\mathrm{C}}(d)=1/(1+(d/\delta)^2)$. This is a controlled comparison in
which only the temporal weighting function is changed. Relative to vanilla attention, Gaussian weighting
improves detection and anticipation by 2.0 and 3.3 percentage points,
respectively. Exponential and Cauchy weighting also outperform vanilla
attention, confirming that temporal modulation is generally useful, but
remain below Gaussian weighting. Specifically, Gaussian exceeds the
exponential kernel by 0.9/1.3 points and the Cauchy kernel by 0.7/1.9 points
in detection/anticipation. The heavier tails retain more distant observations:
this can preserve context, as reflected by the competitive detection results,
but can also introduce less relevant evidence for future-action prediction.
Under this controlled setting, Gaussian weighting provides the most effective
empirical trade-off between local emphasis and long-range context.

\begin{table}[t]
\caption{Ablation study of the future-distribution consistency loss on the
THUMOS'14 test set (mAP, \%). Setting $\lambda_2=0$ removes
$L_{\mathrm{fdc}}$.}
\label{tab:fdc_loss}
\centering
\footnotesize
\setlength{\tabcolsep}{3pt}
\begin{tabular}{@{}lccc@{}}
\toprule
Setting & $\lambda_2$ & Detection & Anticipation \\
\midrule
Without future-distribution consistency loss & 0 & 59.9 & 42.1 \\
Full model & 0.7 & \textbf{71.8} & \textbf{62.6} \\
\bottomrule
\end{tabular}
\end{table}

\textbf{Future-Distribution Consistency Loss.}
Table~\ref{tab:fdc_loss} isolates the contribution of the
future-distribution consistency loss by comparing the full model with an
otherwise identical variant trained with $\lambda_2=0$. Without
$L_{\mathrm{fdc}}$, detection and anticipation achieve 59.9\% and 42.1\%
mAP, respectively. Adding $L_{\mathrm{fdc}}$ improves these results to
71.8\% and 62.6\%, corresponding to gains of 11.9 and 20.5 percentage
points. The larger gain in anticipation is consistent with the loss directly
aligning two future-action distributions. These results demonstrate the
task-level utility of distribution-level agreement.
\begin{figure}[htbp]
  \centering
    \includegraphics[width=1.0\linewidth]{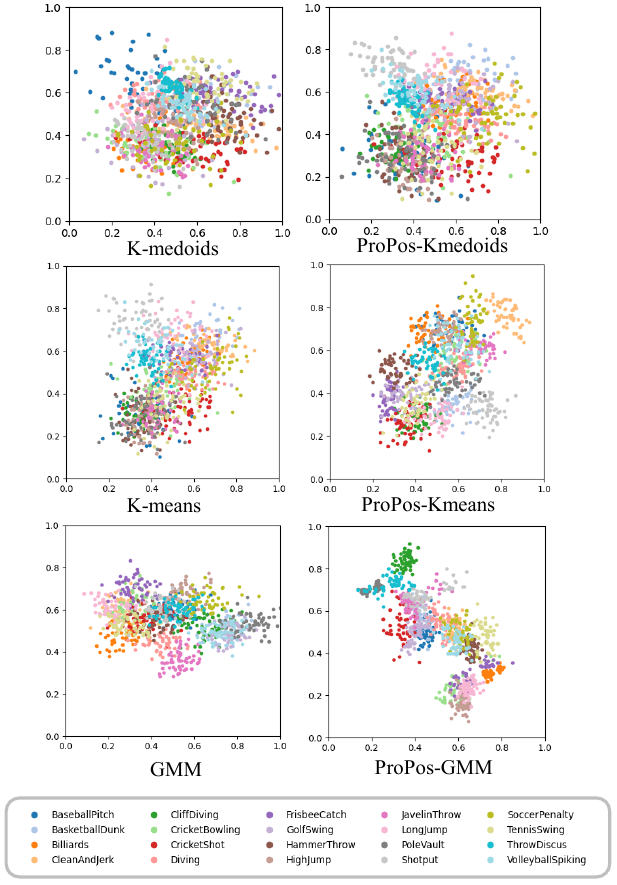}

   \caption{t-SNE visualization of action-feature discriminability under different representative anchor selection strategies. Each point denotes one video frame, and different colours represent the action categories in THUMOS'14.}
   \label{fig2}
\end{figure}
\vspace{-3pt}

\textbf{Representative Anchor Selection Strategy.} To evaluate representative anchor selection in RSMC, we compare different selection strategies in Table~\ref{tab:anchor_selection}. Since APL and CTI depend on the RSMC design, we do not remove the module; instead, we replace ProPos-GMM selection with uniform sampling and stratified random sampling. For stratified random sampling, each video is split into $K$ temporal segments and one frame is randomly drawn from each segment. Table~\ref{tab:anchor_selection} shows that both sampling baselines consistently lag behind ProPos-GMM across various $K$, demonstrating its downstream utility for memory compression. We further evaluate clustering schemes: directly applying K-means/K-medoids/GMM on raw features is inferior to clustering after ProPos-based feature refinement. Overall, ProPos-GMM performs best and peaks at $K=4$. Consistently, the t-SNE results in Fig. 4 exhibit clearer inter-class separation and tighter intra-class clusters, indicating stronger discriminability and providing a better basis for subsequent state compression. 


\begin{table}[h]
\centering
\caption{Ablation study of representative anchor selection strategies (mAP, \%). SRS denotes stratified random sampling.}
\label{tab:anchor_selection}
\footnotesize
\setlength{\tabcolsep}{3.5pt}
\begin{tabular}{cccccccc}
\toprule
\multirow{2}{*}{Method} & \multicolumn{7}{c}{Selected Frame Number} \\ 
\cmidrule{2-8} 
 & 2 & 3 & 4 & 5 & 6&7&8 \\ 
\midrule
Uniform Sampling & 44.1 & 50.3 & 54.9& 56.8 & 55.2&54.1& 52.9\\ 
SRS & 42.0 & 46.6 & 50.1& 54.3 & 56.2&55.6& 52.2\\ 

K-means & 41.9 & 47.4 & 52.0 & 49.3& 48.7&48.0&47.4\\ 
K-medoids & 42.1 & 48.2 & 50.6 & 53.7& 52.9&49.3&48.8\\ 

GMM & 49.9 & 52.2& 57.1 & 60.4& 58.9& 57.2&56.8\\ 

ProPos--K-means & 52.6 & 56.1& 60.8 & 63.5 & 62.0&60.1&59.6\\ 
ProPos--K-medoids & 54.9 & 57.6 & 59.8 & 64.3& 61.6&59.1&58.6\\ 

ProPos-GMM & 61.1 & 65.7 & \textbf{71.8} & 69.1 & 67.3& 65.1&63.9\\ 
\bottomrule
\end{tabular}
\end{table}

\textbf{Within-Component Anchor Selection Criterion.}
We further isolate the effect of the distance used to map each Gaussian
component to an observed anchor frame. Under the same ProPos-GMM configuration,
we compare cosine distance, Mahalanobis distance, maximum component posterior,
and the Euclidean criterion in Eq.~\eqref{eq:euclidean_anchor}. For
Mahalanobis selection, the covariance is regularized as
$\boldsymbol{\Sigma}_k+10^{-6}\mathbf{I}$ using the default covariance
regularization for numerical stability. Maximum posterior selection chooses
$r_k=\arg\max_{i\in\mathcal{I}_k}p(k\mid\mathbf{f}'_i)$. 
\begin{table}[t]
\centering
\caption{Ablation of the within-component anchor selection criterion on the
THUMOS'14 test set. Results are reported in mAP (\%).}
\label{tab:distance_metric}
\footnotesize
\setlength{\tabcolsep}{4pt}
\begin{tabular}{@{}lcc@{}}
\toprule
Anchor criterion & Detection & Anticipation \\
\midrule
Cosine distance      & 71.0 & 61.3 \\
Mahalanobis distance & 69.6 & 60.7 \\
Maximum posterior    & 70.5 & 61.4 \\
Euclidean distance   & 71.8 & 62.6 \\
\bottomrule
\end{tabular}
\end{table}
As shown in Table~\ref{tab:distance_metric}, Euclidean selection achieves the
best detection and anticipation performance, reaching 71.8\% and 62.6\% mAP,
respectively. It improves detection by 0.8, 2.2, and 1.3 percentage points
over cosine distance, Mahalanobis distance, and maximum posterior,
respectively; the corresponding anticipation gains are 1.3, 1.9, and 1.2
points. Euclidean selection directly minimizes the feature-space displacement
between a component mean and its observed representative. In contrast, cosine
distance discards feature-magnitude information, Mahalanobis distance depends
on the covariance estimated from the finite observed buffer, and maximum
posterior favours assignment confidence relative to competing components
rather than proximity to the component center. The comparison therefore
supports Euclidean distance as an empirically effective criterion for anchor selection. 


\begin{table}[t]
\caption{Controlled decomposition of representative-anchor selection and
memory-state construction on the THUMOS'14 test set (mAP, \%). Anchor only denotes the setting without TWA.}
\label{tab:rsmc_decomposition}
\centering
\footnotesize
\setlength{\tabcolsep}{4pt}
\begin{tabular}{@{}llcc@{}}
\toprule
Feature/anchor stage & State construction & Detection & Anticipation \\
\midrule
Uniform sampling & Anchor only & 51.9 & 39.6 \\
Uniform sampling & TWA & 54.9 & 40.7 \\
\midrule
GMM & Anchor only & 53.5 & 39.9 \\
GMM & TWA & 57.1 & 43.0 \\
\midrule
ProPos--GMM & Anchor only & 67.9 & 54.1 \\
ProPos--GMM & Vanilla attention & 69.8 & 59.3 \\
ProPos--GMM & TWA & 71.8 & 62.6 \\
\bottomrule
\end{tabular}
\end{table}

\textbf{Controlled Decomposition of ProPos and RSMC.}
The controlled study in Table~\ref{tab:rsmc_decomposition} distinguishes the
benefit of the adopted ProPos representation from that of the proposed
task-specific state construction. Under the common anchor-only
construction, replacing GMM with ProPos--GMM improves
detection from 53.5\% to 67.9\% and anticipation from 39.9\% to 54.1\%.
We attribute these 14.4/14.2-point gains to the stronger feature space
provided by the adopted ProPos method, rather than claiming them as evidence
of a new clustering algorithm. Holding ProPos--GMM fixed, vanilla attention
improves over anchor only by 1.9/5.2 points, and the proposed TWA further
improves over vanilla attention by 2.0/3.3 points, reaching
71.8\%/62.6\%. Thus, the complete improvement is not obtained by applying
ProPos alone: context aggregation, particularly its temporally weighted
form, contributes additional task-level gains.

The same table further tests whether TWA is tied to a particular selector.
Compared with anchor only, TWA improves detection/anticipation by 3.0/1.1
points with uniform sampling, 3.6/3.1 points with GMM, and 3.9/8.5 points
with ProPos--GMM. The gains across a non-clustering sampler, raw feature
clustering, and ProPos-refined clustering show that TWA consistently benefits
all evaluated anchor selectors and complements, rather than merely inherits,
the benefit of ProPos. Together with the subsequent APL and CTI ablations, which hold
the ProPos--GMM selector fixed, these results separate the contribution of
the adopted anchor initialization from those of state construction,
relational modelling, and cross-temporal interaction.

\textbf{Label-Guided Prototype Baselines and Anchor Diagnostics.}
We compare the proposed label-agnostic selection with two
training-label-guided prototype baselines on THUMOS'14. This comparison
examines whether frame-level training annotations provide a better basis for
anchor selection and whether the label-agnostic selectors are dominated by
background observations. These baselines use
the same setting; only the historical-anchor selection rule is
changed. Let
$\mathbf{z}_i=\operatorname{Norm}(\phi_{\mathrm{ProPos}}(\mathbf{f}_i))$
denote the normalized ProPos feature of a unique training timestamp. For
action class $c$, its fixed prototype is computed using only labelled
training features:
\begin{equation}
\mathbf{q}_c =
\operatorname{Norm}\left(
\frac{1}{N_c}\sum_{i:y_{i,c}=1}\mathbf{z}_i
\right),
\label{eq:action_prototype}
\end{equation}
where $y_{i,c}$ is the frame-level training annotation, $N_c$ is the number
of training timestamps annotated with class $c$, and
$\operatorname{Norm}(\cdot)$ denotes $L_2$ normalization. A timestamp with
overlapping actions contributes to every corresponding action prototype. The
background prototype is
\begin{equation}
\mathbf{q}_{\mathrm{bg}} =
\operatorname{Norm}\left(
\frac{1}{N_{\mathrm{bg}}}
\sum_{i:\sum_c y_{i,c}=0}\mathbf{z}_i
\right).
\label{eq:background_prototype}
\end{equation}
Each unique training timestamp is counted once when the prototypes are
constructed, rather than being repeatedly counted through overlapping
training buffers.

During both model training and inference, the ground-truth label of a
candidate anchor is not used. Given an observed buffer, each normalized frame
feature is compared with the fixed prototypes by cosine similarity,
$s_{i,c}=\mathbf{z}_i^\top\mathbf{q}_c$, and assigned to its most similar
candidate prototype. For each non-empty prototype, the observed frame with
the highest similarity is used as its representative candidate. The $K$
prototypes with the greatest numbers of assigned frames are retained. If
fewer than $K$ prototypes are active, the remaining anchors are filled by
unselected frames in descending order of their maximum prototype similarity.
The Action Prototype variant uses only action prototypes as anchor candidates,
whereas the Action+BG Prototype variant additionally allows the background
prototype to contribute an anchor. The selected items are always actual
observed frame features rather than the prototype vectors themselves.

We further diagnose the selected historical anchors with two metrics. Let
$r_{t,k}$ be the timestamp of the $k$-th selected historical anchor in the
buffer ending at $t$, and let $b(r_{t,k})=1$ if that timestamp has at least
one ground-truth foreground-action annotation. For $T$ evaluated buffers,
the foreground anchor ratio is
\begin{equation}
\mathrm{FAR} =
\frac{1}{TK}\sum_{t=1}^{T}\sum_{k=1}^{K} b(r_{t,k}).
\label{eq:far}
\end{equation}
FAR measures foreground purity but is not itself an optimization objective,
because background and transition context may also be useful. Let
$\mathcal{Y}_t$ denote the set of foreground action categories occurring in
buffer $t$ and let $\mathcal{A}_t$ denote the set of ground-truth foreground
categories attached to its selected historical anchors. Action coverage is
\begin{equation}
\mathrm{AC} =
\frac{1}{|\mathcal{T}_{+}|}
\sum_{t\in\mathcal{T}_{+}}
\frac{|\mathcal{A}_t\cap\mathcal{Y}_t|}{|\mathcal{Y}_t|},
\qquad
\mathcal{T}_{+}=\{t:|\mathcal{Y}_t|>0\}.
\label{eq:action_coverage}
\end{equation}
Repeated anchors from the same action category count once in
$\mathcal{A}_t$. The current frame, which is appended separately in RSMC, is
excluded from both FAR and AC. Test annotations are accessed only after anchor
selection to calculate these diagnostic metrics; they are never used to
construct prototypes, select anchors, or produce task predictions.

Table~\ref{tab:prototype_selection} compares the label-agnostic and
label-guided anchor selectors in terms of anchor characteristics and
downstream task performance.
\begin{table}[t]
\caption{Comparison of anchor selection methods on THUMOS'14 with $K=4$.
Sup. indicates whether training labels are used to construct fixed prototypes.
FAR and AC denote foreground anchor ratio and action coverage, respectively,
and are reported in percentage. Test annotations are used only to calculate
FAR and AC after anchor selection. Det. and Ant. are reported in mAP (\%).}
\label{tab:prototype_selection}
\centering
\footnotesize
\setlength{\tabcolsep}{2.5pt}
\begin{tabular}{@{}lccccc@{}}
\toprule
Method & Sup. & FAR & AC & Det. & Ant. \\
& & & $\uparrow$ & $\uparrow$ & $\uparrow$ \\
\midrule
GMM & No & 68.5 & 55.2 & 57.1 & 43.0 \\
ProPos--GMM & No & 75.8 & \textbf{70.3} & \textbf{71.8} & \textbf{62.6} \\
ProPos--Act. Proto. & Yes & \textbf{89.4} & 67.5 & 68.1 & 55.3 \\
ProPos--Act.+BG Proto. & Yes & 79.8 & 66.2 & 69.5 & 57.7 \\
\bottomrule
\end{tabular}
\end{table}
As shown in Table~\ref{tab:prototype_selection}, we first compare GMM and
ProPos--GMM to assess the effect of ProPos feature refinement. ProPos--GMM
improves over GMM by 7.3 points in FAR and 15.1 points in AC,
along with gains of 14.7 and 19.6 mAP in detection and anticipation,
respectively. This comparison quantifies the contribution of the adopted
ProPos feature refinement rather than constituting evidence for a novel
clustering algorithm.

We next examine whether stronger foreground bias necessarily yields more
useful anchors. The action-only prototype selector obtains the highest FAR of
89.4\%,
confirming that training-label guidance can impose a stronger foreground bias.
However, its AC is 2.8 points lower than that of ProPos--GMM, while detection
and anticipation decrease by 3.7 and 7.3 mAP. Thus, selecting a larger
proportion of foreground anchors does not by itself produce a more useful
online memory: fixed action centers can repeatedly favour class-typical
observations while underrepresenting the diverse modes and transitions present
in a particular observed buffer. Consistent with the diagnostic role of FAR
described above, the highest foreground purity does not yield the best task
performance.

Finally, we evaluate whether background context is useful. Adding a background
prototype reduces FAR from 89.4\% to 79.8\%, but improves
detection from 68.1\% to 69.5\% and anticipation from 55.3\% to 57.7\%.
This comparison indicates that background and action--background context can
provide complementary information, although foreground purity is reduced.
Compared with this stronger supervised variant, ProPos--GMM still improves AC
by 4.1 points and detection/anticipation by 2.3/4.9 mAP, despite using no
frame-level labels for anchor selection.


Taken together, the results support
our interpretation of RSMC as label-agnostic, window-adaptive representative
compression rather than foreground localization: its advantage arises from
balancing foreground evidence, action-category coverage, and contextual
diversity, rather than maximizing FAR alone.

\textbf{Relational Modelling Configuration.}
We first investigate whether the benefit of APL stems from graph message propagation and pair-specific relation modelling, rather than merely from representative state features or additional nonlinear capacity. All variants in Table~\ref{tab:apl_relation} use the same settings, differing only in how the graph-level representation is constructed. In mean state pooling, the representative states are averaged and linearly projected to replace the graph-level representation, without graph construction or message propagation. Nonlinear state pooling applies the same two-layer MLP independently to every representative state and then averages the transformed state features:
\begin{equation}
\mathbf{F}_{\mathrm{MLP}}
=\frac{1}{K+1}\sum_{i=1}^{K+1}
\mathbf{W}_2\phi(\mathbf{W}_1\mathbf{S}_i+\mathbf{b}_1)+\mathbf{b}_2,
\label{eq:mlp_pooling}
\end{equation}
where $\phi(\cdot)$ denotes the activation function. This variant produces a latent representation with the same dimensionality as $\mathbf{F}_g$ but introduces neither pairwise edge encoding nor inter-state message propagation. Shared-edge gated GCN retains the fully connected graph and Gated GCN but assigns a single learnable edge vector shared by all connected state pairs, thereby removing pair-specific edge encoding. The complete APL uses the proposed pair-specific multi-dimensional edges to regulate Gated GCN message propagation. These variants therefore progressively evaluate the effects of nonlinear transformation, graph propagation, and pair-specific relation modelling.
\begin{table}[t]
\caption{Ablation study of relational modelling configurations in APL on the THUMOS'14 test set (mAP, \%).}
\label{tab:apl_relation}
\centering
\footnotesize
\setlength{\tabcolsep}{3pt}
\begin{tabular}{@{}lcc@{}}
\toprule
Relation Modelling Configuration & Detection & Anticipation \\
\midrule
Mean State Pooling & 33.7 & 12.1 \\
Nonlinear State Pooling (MLP) & 41.4 & 26.5 \\
Shared-edge Gated GCN & 52.3 & 41.9 \\
APL with Multi-dimensional Edges & \textbf{71.8} & \textbf{62.6} \\
\bottomrule
\end{tabular}
\end{table}

Mean state pooling obtains 33.7\% detection and 12.1\% anticipation mAP, showing that directly aggregating the representative states provides only a weak substitute for the graph representation. Nonlinear state pooling improves these results to 41.4\% and 26.5\%, yielding gains of 7.7 and 14.4 percentage points, respectively. This improvement demonstrates the benefit of additional nonlinear capacity, although the states remain independently processed. Adding Gated GCN message propagation with a shared edge further increases detection and anticipation to 52.3\% and 41.9\%, outperforming nonlinear pooling by 10.9 and 15.4 percentage points. This result shows that inter-state message propagation provides substantial benefits beyond independent state transformation and pooling. The complete APL achieves 71.8\% detection and 62.6\% anticipation mAP, improving over the shared-edge graph by 19.5 and 20.7 percentage points. It also outperforms nonlinear state pooling by 30.4 and 36.1 percentage points and mean state pooling by 38.1 and 50.5 percentage points, respectively. Because all other components remain unchanged, these results demonstrate the progressive contributions of nonlinear state transformation, graph message propagation, and pair-specific multi-dimensional relation modelling. In particular, the large improvement over nonlinear state pooling confirms that APL's effectiveness cannot be attributed solely to additional nonlinear transformation capacity.

\begin{table}[t]
\caption{Linear probing of current- and future-action information on THUMOS'14 (mAP, \%). For each frozen representation, two independent linear classifiers with identical architectures are trained on the training split using the current-action label and the future-action label at $\tau=1$ s, respectively, and evaluated on the test split.}
\label{tab:future_probe}
\centering
\footnotesize
\setlength{\tabcolsep}{4pt}
\begin{tabular}{@{}lcc@{}}
\toprule
Representation & Current-Action & Future-Action \\
& mAP (\%) & mAP (\%) \\
\midrule
Present cue $\mathbf{F}_c$ & \textbf{50.9} & 8.6 \\
RSMC state mean $\overline{\mathbf{S}}$ & 26.1 & 15.3 \\
Future-oriented graph cue $\mathbf{F}_g$ & 37.7 & 43.9 \\
CTI-refined future representation $\mathbf{F}'_g$ & 41.0 & \textbf{50.8} \\
\bottomrule
\end{tabular}
\end{table}

\textbf{Future-Action Information Probing.}
We use linear probing to ask a question: as the representation progresses from the present cue to the graph-derived cue and its CTI-refined form, does linearly accessible information shift from the ongoing action toward the future action? To answer this question, we freeze the learned feature encoders and train two independent, newly initialized linear probes for each representation, one for the current action and the other for the future action. The probes operate directly on the frozen representations, without the original prediction heads or any additional refinement. Thus, this experiment is diagnostic rather than an alternative inference configuration; the complete model still uses $\mathbf{F}'_c$ for detection and $\mathbf{F}'_g$ for anticipation.

The RSMC state mean $\overline{\mathbf{S}}$ provides a relation-free reference for determining whether future-action information comes merely from aggregating the observed history or from graph reasoning. Although it is computed using the same averaging operation as mean state pooling in Table~\ref{tab:apl_relation}, the two experiments serve different purposes. Table~\ref{tab:apl_relation} tests an end-to-end replacement of APL while retaining CTI and the original prediction heads, whereas Table~\ref{tab:future_probe} measures the information directly accessible from each frozen representation through linear classifiers. Keeping this distinction clear allows the probing results to be interpreted as representational evidence rather than another model ablation.

Table~\ref{tab:future_probe} reveals a clear progression. First, the present cue $\mathbf{F}_c$ is strongly biased toward the ongoing action: it achieves 50.9\% current-action mAP but only 8.6\% future-action mAP. Second, averaging the RSMC states raises future-action mAP to 15.3\%, showing that the observed history contains some future-relevant context, but its current-action mAP falls to 26.1\%, indicating that relation-free temporal aggregation weakens current-action discrimination. Third, applying graph reasoning produces $\mathbf{F}_g$ and increases future-action mAP from 15.3\% to 43.9\%. Unlike $\mathbf{F}_c$, $\mathbf{F}_g$ is more predictive of the future action than of the current action (43.9\% versus 37.7\%), demonstrating that the graph representation changes the information emphasis rather than simply preserving a generic pooled summary.

Finally, CTI raises future-action mAP by a further 6.9 percentage points, from 43.9\% for $\mathbf{F}_g$ to 50.8\% for $\mathbf{F}'_g$, while current-action mAP increases by only 3.3 points, from 37.7\% to 41.0\%. This asymmetric gain indicates that CTI primarily strengthens future-relevant information while retaining complementary current-action context. Taken together, the progression from $\overline{\mathbf{S}}$ to $\mathbf{F}_g$ and then $\mathbf{F}'_g$ supports our operational interpretation of $\mathbf{F}_g$ as a future-oriented graph cue rather than a generic pooled latent representation. 

\begin{figure}[htbp]
  \centering
    \includegraphics[width=0.9\linewidth]{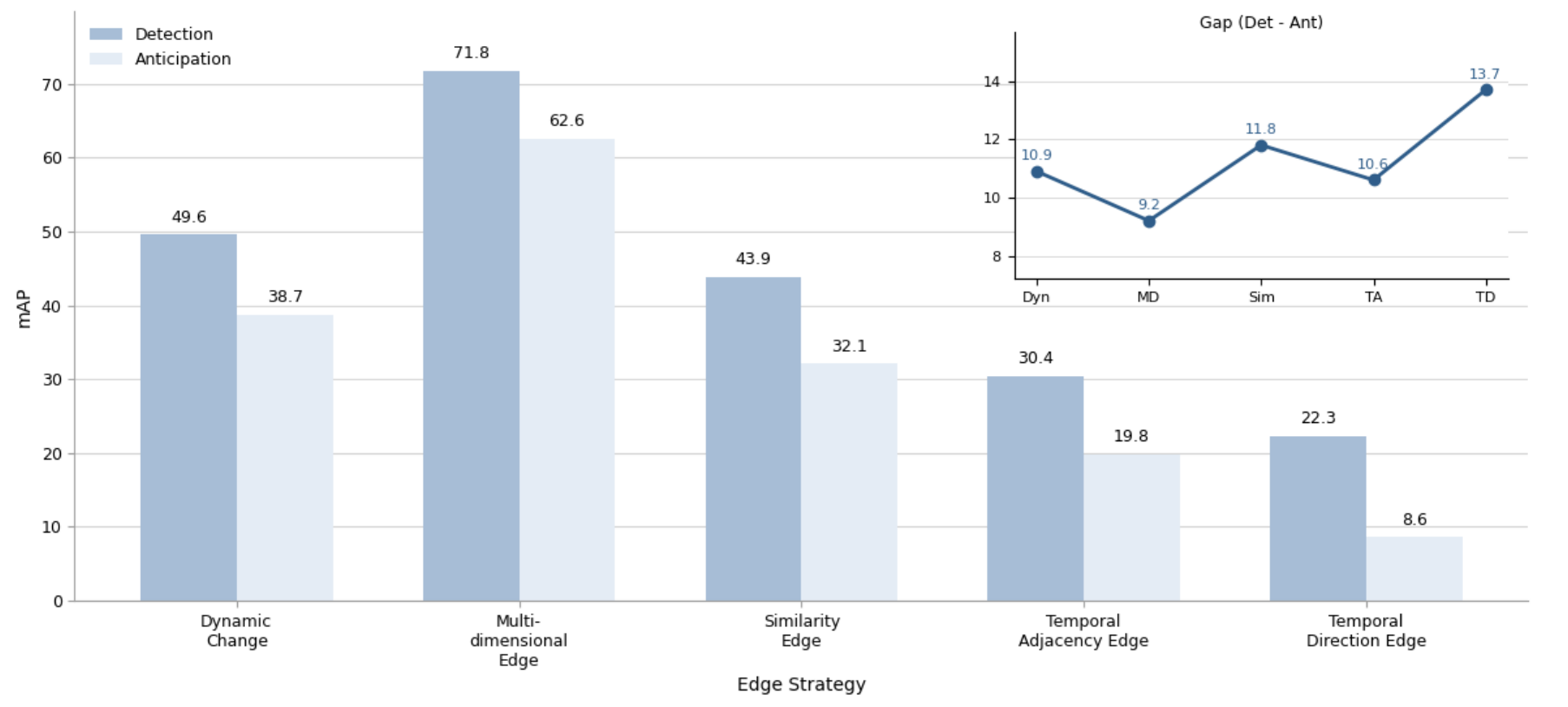}

   \caption{Ablation study of inter-state edge strategies. All variants use the same representative states produced by RSMC and differ only in their graph relation encoding. Dyn, MD, Sim, TA and TD denote dynamic change, multi-dimensional edge, similarity edge, temporal adjacency edge and temporal direction edge, respectively.}
   \label{fig:edge_strategy}
\end{figure}
\textbf{Edge Strategy.}
Having isolated the contributions of graph propagation and pair-specific relation modelling, we further compare alternative edge encodings in Fig.~\ref{fig:edge_strategy}. All edge-strategy variants use the same representative states produced by RSMC and the same Gated GCN, differing only in how graph edges are defined. We replace the proposed multi-dimensional edges with four single-type relation encodings: Temporal Direction, Temporal Adjacency, Similarity, and Dynamic Change. Here, lowercase $e_{i,j}$ denotes a predefined scalar edge weight, distinguishing it from the proposed multi-dimensional edge vector $\mathbf{E}_{i,j}$. For Temporal Direction, we set $e_{i,j}=1$ if the anchor frame of state $\mathbf{S}_j$ occurs after that of $\mathbf{S}_i$, and $e_{i,j}=0$ otherwise. Temporal Adjacency assigns larger weights to temporally closer states, where $w_{i,j}=\frac{1}{|i-j|}$ and $e_{i,j}=\frac{w_{i,j}}{\sum_{k\in\mathcal{N}(i)} w_{i,k}}$, with $\mathcal{N}(i)$ denoting the neighborhood of node $i$. Similarity encodes edges using cosine similarity, i.e., $e_{i,j}=\cos(\mathbf{S}_i,\mathbf{S}_j)=\frac{\mathbf{S}_i^\top \mathbf{S}_j}{\|\mathbf{S}_i\|_2\,\|\mathbf{S}_j\|_2}$. For Dynamic Change, we define $e_{i,j}=\|\mathbf{S}_i-\mathbf{S}_j\|_2$ to measure the magnitude of feature variation between two representative states. Among the four predefined scalar relations, Dynamic Change performs best, achieving 49.6\% detection and 38.7\% anticipation. Nevertheless, the proposed multi-dimensional edge substantially improves these results to 71.8\% and 62.6\%, corresponding to gains of 22.2 and 23.9 percentage points, respectively. These results indicate that no single predefined relation, including temporal direction, temporal adjacency, feature similarity, or feature variation, is sufficient to represent the evolving dependencies among the observed states. Instead, jointly learned multi-dimensional edges provide a more expressive basis for graph-level action-dynamics modelling. Together with Table~\ref{tab:apl_relation}, these controlled experiments demonstrate that the graph-level action-dynamics representation benefits from both relational message propagation and pair-specific multi-dimensional edge encoding. They provide task-level evidence for representing action dynamics, but do not imply that an individual edge channel corresponds to an explicitly annotated action phase, transition type, or causal relation.

\begin{table}
\caption{Ablation study of feature-refinement paths in CTI on the
THUMOS'14 test set (mAP, \%). Here,
$\mathbf{X}\!\rightarrow\!\mathbf{Y}$ indicates that cue $\mathbf{X}$ is
used as contextual information to refine target representation $\mathbf{Y}$
through cross-attention, while
$\{\mathbf{X}_1,\mathbf{X}_2\}\!\rightarrow\!\mathbf{Y}$ indicates that both
cues are used jointly. $\mathbf{F}_p$, $\mathbf{F}_c$, and $\mathbf{F}_g$
denote the past cue, present cue, and future-oriented graph cue, respectively; a
prime denotes a refined representation. ``None'' indicates that the
corresponding cross-temporal update is not applied.}
  \centering
  \footnotesize
  \setlength{\tabcolsep}{2.5pt}
  \begin{tabular}{@{}lccll@{}}
    \toprule
    No. & \shortstack{Present-repr.\\update}
        & \shortstack{Future-repr.\\update}
        & Det. & Ant. \\
    \midrule
    (1) & None & None & 49.4 & 44.3 \\
    (2) & $\mathbf{F}_p\!\rightarrow\!\mathbf{F}'_c$ & None & 63.5 & 44.9 \\
    (3) & None & $\mathbf{F}_p\!\rightarrow\!\mathbf{F}'_g$ & 49.6 & 52.0 \\
    (4) & None & $\mathbf{F}_c\!\rightarrow\!\mathbf{F}'_g$ & 49.9 & 55.3 \\
    (5) & $\mathbf{F}_g\!\rightarrow\!\mathbf{F}'_c$ & None & 59.6 & 45.4 \\
    (6) & $\{\mathbf{F}_p,\mathbf{F}_g\}\!\rightarrow\!\mathbf{F}'_c$
        & $\{\mathbf{F}_p,\mathbf{F}'_c\}\!\rightarrow\!\mathbf{F}'_g$
        & 71.8 & 62.6 \\
    \bottomrule
  \end{tabular}
  
  \label{tab:cti}
  
\end{table}

\textbf{Cross-Temporal Interaction.} Table~\ref{tab:cti} analyses how individual feature-refinement paths in CTI affect online action detection and anticipation. Without cross-temporal interaction (Case (1)), the model uses the unrefined present and graph representations and achieves 49.4\% detection and 44.3\% anticipation. Using the past cue to refine only the present representation (Case (2)) improves detection to 63.5\%, whereas using it to refine only the future-oriented representation (Case (3)) improves anticipation to 52.0\%. The direction-specific comparison between the present and graph cues further demonstrates their complementary roles. Using the present cue to update the future-oriented representation (Case (4)) raises anticipation from 44.3\% to 55.3\%, while using the graph cue to update the present representation (Case (5)) raises detection from 49.4\% to 59.6\%. The complete two-stage configuration (Case (6)) first refines $\mathbf{F}_c$ using $\mathbf{F}_p$ and $\mathbf{F}_g$ and then refines $\mathbf{F}_g$ using $\mathbf{F}_p$ and $\mathbf{F}'_c$, achieving the best results of 71.8\% detection and 62.6\% anticipation. These results support complementary cross-temporal feature sharing and task-specific representation refinement.
\subsection{Comparison with Current Representative Methods}
\begin{table}
\caption{Comparison with prior work on EPIC-Kitchens-100 for action anticipation.}
\label{tab:epic_anticipation}
  \centering
  \begin{tabular}{ccccc}
    \toprule
    Method & Modality&Verb&Noun&Action \\
    \midrule
    RULSTM\cite{furnari2019would}&RGB&27.5& 29.0& 13.3\\
    AVT\cite{girdhar2021anticipative} &RGB&30.2& 31.7& 14.9\\
    TeSTra\cite{zhao2022real}& RGB&26.8& 36.2& 17.0 \\
     MeMViT\cite{wu2022memvit}& RGB&32.8&33.2&15.1\\
    MAT\cite{wang2023memory}&RGB& \underline{32.7} & \textbf{39.7}& 18.8 \\
    S-GEAR$-$2B\cite{diko2024semantically} & RGB&\underline{32.7}  &37.9 & \underline{19.6}\\
    CPM\cite{xie2024towards}& RGB& -&-& 17.2\\
    Ours& RGB& \textbf{36.8}& \underline{39.2}&\textbf{19.9}\\
    \midrule
    TeSTra\cite{zhao2022real}& RGB+OF&30.8& 35.8& 17.6 \\
    MAT\cite{wang2023memory}& RGB+OF& \underline{35.0}& \underline{38.8}& 19.5 \\
    S-GEAR$-$2B\cite{diko2024semantically}& RGB+Obj &30.5 &38.4&19.6\\
    S-GEAR$-$4B\cite{diko2024semantically} & RGB+Obj  &30.2  &37.0 & \underline{19.9}\\
    Ours& RGB+OF& \textbf{38.8}&\textbf{42.1} & \textbf{21.4}\\
    \midrule
    RULSTM\cite{furnari2019would}&RGB+OF+Obj&27.8& 30.8& 14.0\\
   AVT+\cite{girdhar2021anticipative}&RGB+OF+Obj&28.2& 32.0& 15.9\\
    CPM\cite{xie2024towards}& RGB+OF+Obj& -&-& 19.4\\
    
    UADT \cite{guo2024uncertainty}& RGB+OF+Obj &\underline{43.5}& \underline{46.6}& \underline{23.0}\\
    
    Ours& RGB+OF+Obj& \textbf{44.9}&\textbf{48.3} & \textbf{24.9}\\

    \bottomrule
  \end{tabular}
  
\end{table}

\begin{table}[htbp]
\caption{Online action detection performance on THUMOS'14 and TVSeries.}
\label{tab:online_detection}
\centering
\resizebox{\linewidth}{!}{%
\begin{tabular}{lcccc}
  \toprule
  \multirow{2}{*}{\textbf{Method}} 
    & \multicolumn{2}{c}{\textbf{THUMOS'14}} 
    & \multicolumn{2}{c}{\textbf{TVSeries}} \\
  \cmidrule(lr){2-3}
  \cmidrule(lr){4-5}
  & \textbf{Kinetics} & \textbf{ANet} 
  & \textbf{Kinetics} & \textbf{ANet} \\
  \midrule
  TRN\cite{xu2019temporal} & 62.1 & 47.2 & 86.2 & 83.7 \\
  OadTR\cite{wang2021oadtr} & 65.2 & 58.3 & 87.2 & 85.41 \\
  Colar\cite{yang2022colar} & 66.9 & 59.4 & 88.1 & 86.0 \\
  LSTR\cite{xu2021long} & 69.5 & 65.3 & 89.1 & 88.1 \\
  GateHub\cite{chen2022gatehub} & 70.7 & 69.1 & 89.6 & 88.4 \\
  TeSTra\cite{zhao2022real} & \underline{71.2} & 68.2 & - & - \\
  MAT\cite{wang2023memory} & 71.6 & \underline{70.4} & \underline{89.7} & \underline{88.6} \\
  HCM\cite{liu2023hcm} & 68.7 & 66.2 & 88.2 & - \\
  ADI-Diff\cite{foo2024action} & 70.8 & - & - & - \\
  ContextDet\cite{wang2024contextdet} & 69.5 & - & - & - \\
  Ours & \textbf{72.1} & \textbf{71.8} & \textbf{90.4} & \textbf{89.8} \\
  \bottomrule
\end{tabular}%
}
\end{table}

\begin{table*}[t]
\centering
\caption{Action anticipation results on THUMOS'14 and TVSeries with different anticipation horizons. mAP is reported for THUMOS'14 and mcAP for TVSeries.  “ANet” and “Kinetics” denote features pretrained on ActivityNet v1.3 and Kinetics-400, respectively. }
\label{tab:anticipation_thumos_tv}
\setlength{\tabcolsep}{7pt}
\begin{tabular}{cccccccccccc}
\toprule
\multirow{2}{*}{Dataset} & \multirow{2}{*}{Method} &\multirow{2}{*}{Pretrained} &
\multicolumn{8}{c}{Anticipation Time Gap} & \multirow{2}{*}{Avg.} \\
& & &0.25s & 0.50s & 0.75s & 1.00s & 1.25s & 1.50s & 1.75s & 2.00s & \\
\midrule
\multirow[c]{15}{*}{THUMOS 14}
& TRN \cite{xu2019temporal}&ANet
& 45.1 & 42.4 & 40.7 & 39.1 & 37.7 & 36.4 & 35.3 & 34.3 & 38.9 \\
& LSTR\cite{xu2021long}&ANet
& - & - & - & - & - & - & - & -  & 50.1\\
& TeSTra\cite{zhao2022real}&ANet
& \underline{64.7} & \underline{61.8} & \underline{58.7} & \underline{55.7} & \underline{53.2} & \underline{51.1} & \underline{49.2} & \underline{47.8}  & 55.3\\
& OadTR \cite{wang2021oadtr}&ANet
& 49.3 & 48.1 & 46.8 & 45.3 & 43.9 & 42.4 & 41.1 & 40.1 & 45.9 \\
& HCM \cite{liu2023hcm}&ANet
& 62.4 & 59.4 & 56.3 & 53.3 & 50.5 & 48.3 & 46.4 & 44.8 & 52.8 \\
& MAT\cite{wang2023memory}&ANet
& - & - & - & - & - & - & - & -  & \underline{57.3}\\
& Ours&ANet
& \textbf{69.2} & \textbf{65.3} & \textbf{62.2} & \textbf{59.2} & \textbf{56.7} & \textbf{54.6} & \textbf{52.7} &\textbf{51.3}  & \textbf{58.9}\\
\cmidrule(lr){2-12}
& LSTR \cite{xu2021long}&Kinetics
& 60.4 & 58.6 & 56.0 & 53.3 & 50.9 & 48.9 & 47.1 & 45.7 & 52.6\\
& OadTR \cite{wang2021oadtr}&Kinetics
& 59.8 & 58.5 & 56.6 & 54.6 & 52.6 & 50.5 & 48.6 & 46.8 & 53.5 \\
& TeSTra\cite{zhao2022real}&Kinetics
& 66.2 & 63.5 & 60.5 & 57.4 & 54.8 & 52.6 & 50.5 & 48.9  & 56.8\\
& HCM\cite{liu2023hcm} &Kinetics
& 64.6 & 61.4 & 57.6 & 54.6 & 51.8 & 49.3 & 46.8 & 45.3 & 54.1 \\
& MAT\cite{wang2023memory}&Kinetics
& - & - & - & - & - & - & - & -  & 58.2\\
& CMeRT\cite{pang2025context}&-
& \underline{69.9} & \underline{66.6} & \underline{63.2} & \underline{60.1} & \underline{57.3} & \underline{54.9} & \underline{52.8} & \underline{50.9}  & \underline{59.5}\\
& Ours&Kinetics
& \textbf{72.4} & \textbf{69.1} & \textbf{65.7} & \textbf{62.6} & \textbf{59.8} & \textbf{57.4}& \textbf{55.3} & \textbf{52.9}  & \textbf{61.9}\\
\midrule
\multirow[c]{9}{*}{TVSeries}
& OadTR\cite{wang2021oadtr}&ANet
& \underline{81.9} & \underline{80.6} & \underline{79.4} & \underline{78.2} & \underline{77.1} & \underline{76.0} & \underline{75.2} & \underline{74.3} & 77.8 \\
& MAT\cite{wang2023memory}&ANet
& - & - & - & - & - & - & - & -  & \underline{81.5}\\
& Ours&ANet
& \textbf{87.7} & \textbf{86.4} & \textbf{85.2} & \textbf{84.0} & \textbf{82.9} & \textbf{81.8} & \textbf{81.0} & \textbf{80.6}  & \textbf{83.7}\\
\cmidrule(lr){2-12}
& TRN \cite{xu2019temporal}&Kinetics
& 79.9 & 78.4 & 77.1 & 75.9 & 74.9 & 73.9 & 73.0 & 72.3  & 75.7\\
& OadTR \cite{wang2021oadtr}&Kinetics
& 84.1 & 82.6 & 81.3 & 80.1 & 78.9 & 77.7 & 76.7 & 75.7 & 79.1 \\
& HCM\cite{liu2023hcm} &Kinetics
& \underline{85.8} & \underline{84.1} & \underline{82.6} & \underline{80.9} & \underline{79.7} & \underline{78.9} & \underline{77.6} & \underline{77.0} & 80.9 \\
& MAT\cite{wang2023memory}&Kinetics
& - & - & - & - & - & - & - & -  & \underline{82.6}\\
& Ours&Kinetics
& \textbf{90.1} & \textbf{88.4} & \textbf{86.9} & \textbf{85.2} & \textbf{84.0} & \textbf{83.2} & \textbf{81.9} & \textbf{81.1}  & \textbf{85.1}\\
\bottomrule
\end{tabular}
\end{table*}

We compare the proposed method with representative online action understanding approaches on EPIC-Kitchens-100, THUMOS'14, TVSeries, and PDMB. Detailed PDMB results are provided in Section IV (Generalizability Evaluation on the PDMB Dataset) of the supplementary material.

\subsubsection{Action Anticipation}

Table~\ref{tab:epic_anticipation} compares our method with representative approaches on EPIC-Kitchens-100. Class-mean top-5 recall for verbs, nouns, and actions is reported under different modality configurations. Our method performs consistently across all settings. With RGB alone, it achieves 36.8\%/39.2\%/19.9\% (verb/noun/action), yielding the highest listed verb and action scores while remaining competitive for nouns. Adding optical flow improves performance to 38.8\%/42.1\%/21.4\%, exceeding the listed two-modality baselines. With RGB+OF+Obj, our method achieves 44.9\%/48.3\%/24.9\%, the highest listed scores under the same modality setting, demonstrating effective multimodal feature learning for action anticipation.

We also evaluate action anticipation on THUMOS'14 and TVSeries at varying time gaps (Table~\ref{tab:anticipation_thumos_tv}). Our method achieves the highest listed results across all horizons and both pretraining settings. With Kinetics features, the average performance is 61.9\% on THUMOS'14 and 85.1\% on TVSeries; with ActivityNet features, it is 58.9\% and 83.7\%, respectively. Performance remains consistently high and declines more slowly than that of the listed baselines as the anticipation horizon increases, indicating robust short- and long-horizon anticipation across datasets and feature sources.

\subsubsection{Action Detection}
Table~\ref{tab:online_detection} shows consistent gains across datasets and
feature settings. SSM obtains 72.1\%/71.8\% on THUMOS'14 and
90.4\%/89.8\% on TVSeries with Kinetics/ActivityNet features, respectively,
supporting the combined benefits of representative states, relational
reasoning, and CTI. EPIC-Kitchens-100 results appear in the supplement.

\textbf{Efficiency.}
Under a unified model-side A100 protocol, SSM uses 3.12 GFLOPs and 63.23M
trainable parameters, trains at 29 ms/iter, and runs at 158.8 FPS. It is the third-fastest method in the comparison; full measurement details appear in the supplement.
\begin{figure}[htbp]
  \centering
    \includegraphics[width=0.8\linewidth]{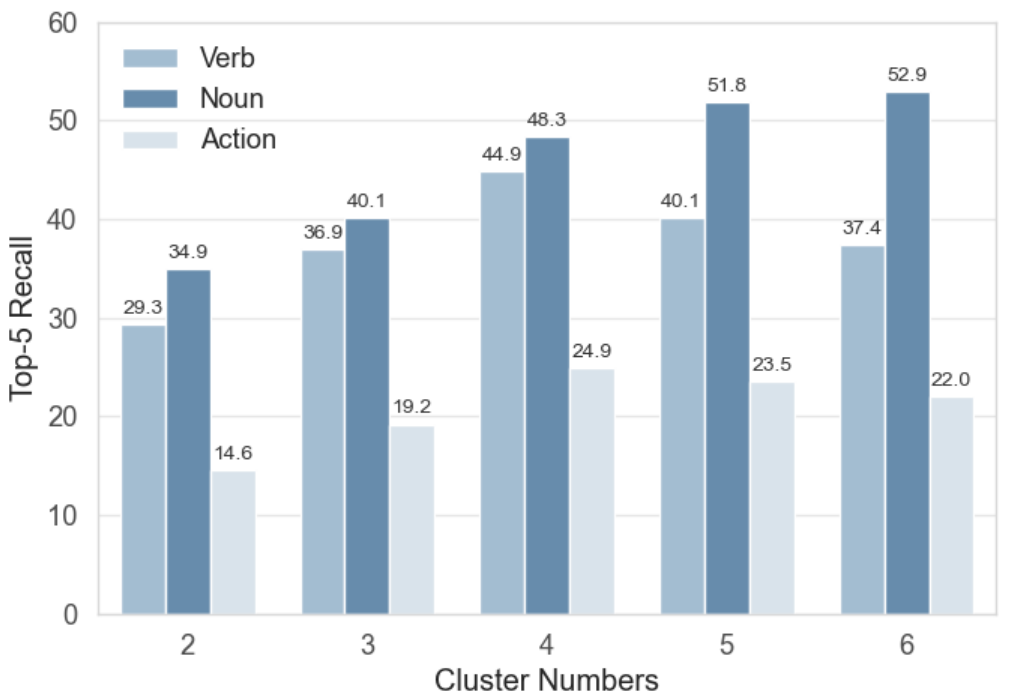}

   \caption{Verb, noun, and action anticipation under different cluster numbers in our representative state-based compression framework.}
   \label{fig:compression_tradeoff_main}
\end{figure}
\begin{figure*}[!b]
  \centering
  \includegraphics[width=0.8\linewidth]{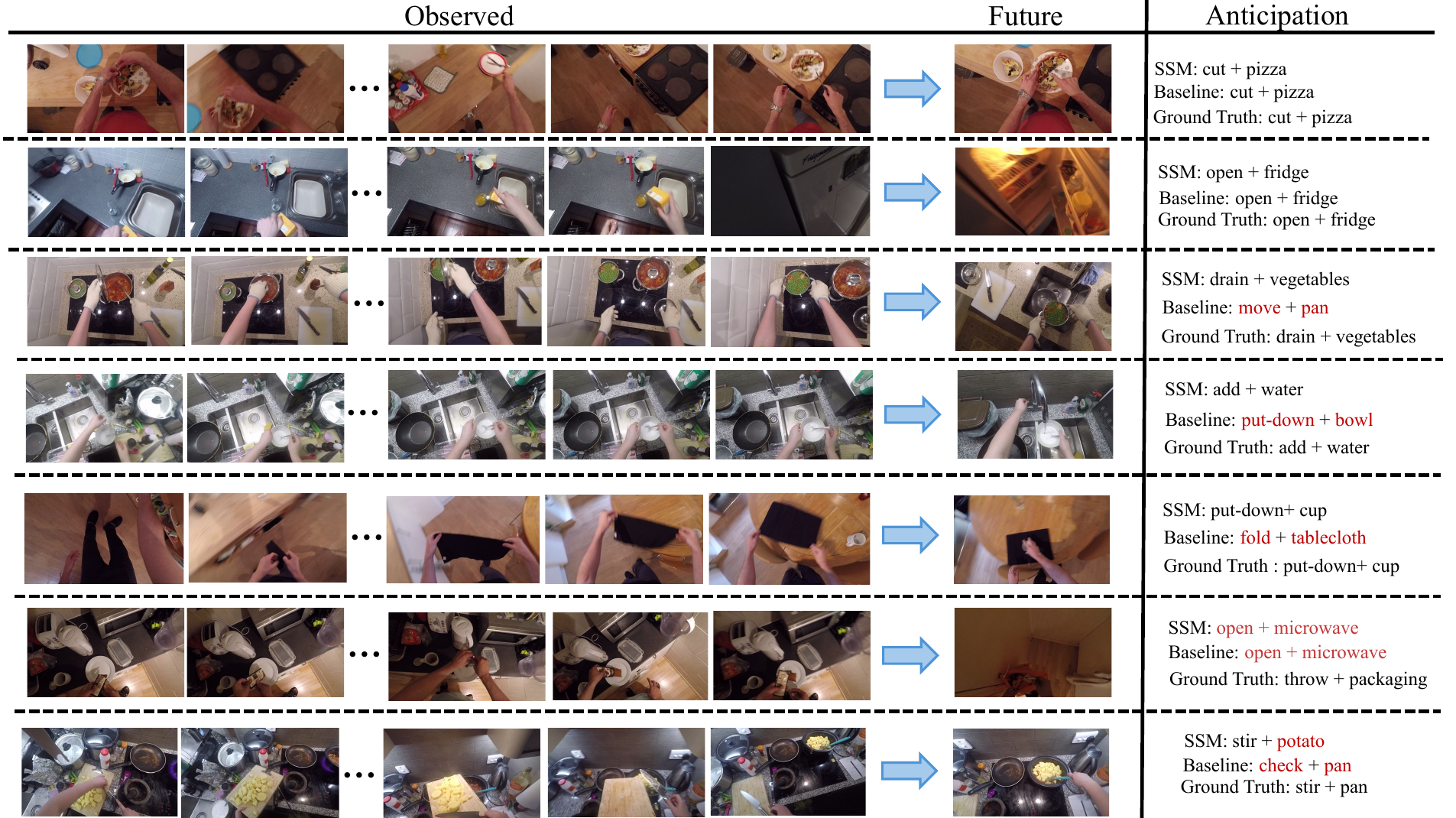}
  \caption{Qualitative anticipation results for SSM and MAT. Incorrect
  anticipations are marked in red.}
  \label{fig:qualitative_anticipation_main}
\end{figure*}
\begin{figure*}[htbp]
  \centering
  \includegraphics[width=0.8\linewidth]{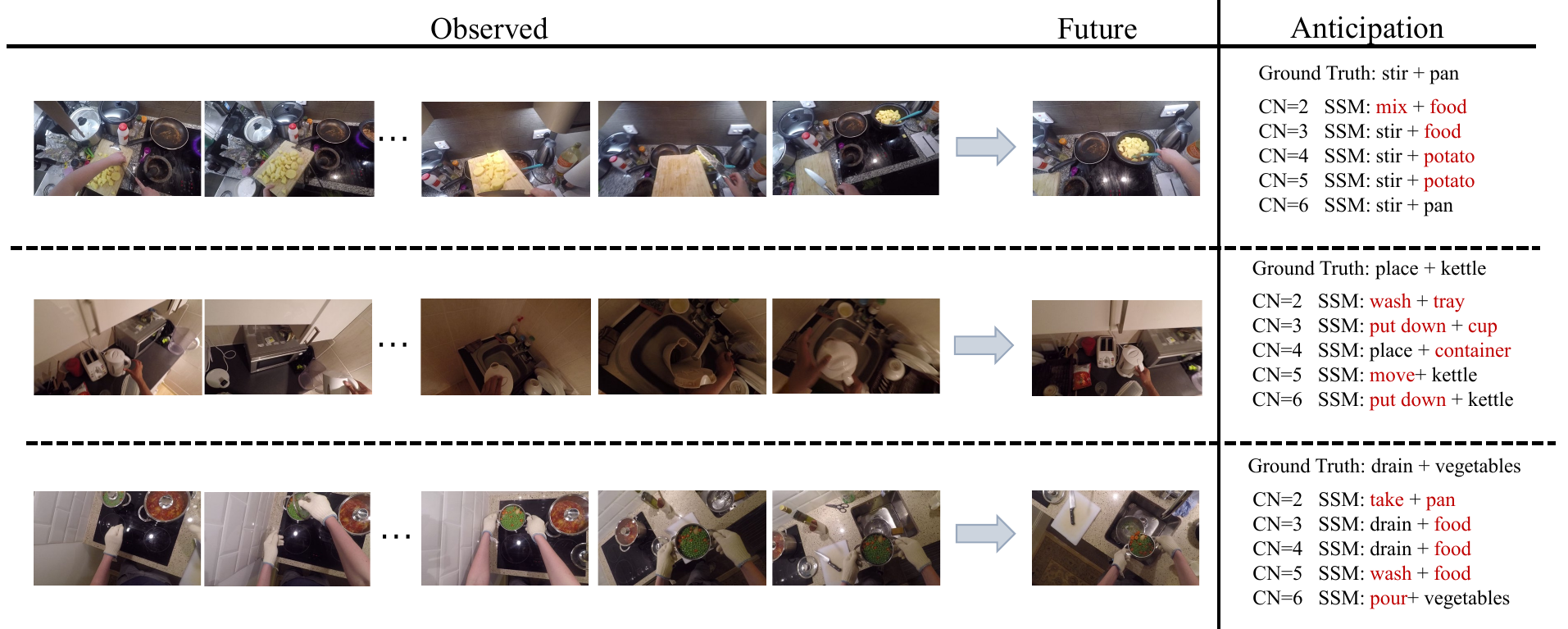}
  \caption{Example-level verb, noun, and action anticipation results under
  different cluster numbers $K$. Finer state granularity often recovers object
  detail, whereas excessive states do not consistently improve verb or action
  anticipation.}
  \label{fig:compression_tradeoff_examples_main}
\end{figure*}
\subsection{Qualitative Analysis}
Fig.~\ref{fig:qualitative_anticipation_main} qualitatively compares the
anticipation performance of SSM and MAT~\cite{wang2023memory} (baseline) on
EPIC-Kitchens-100. Both methods succeed when the observed pattern is clear
(Rows 1--2), whereas SSM better captures less immediate dependencies in
Rows 3--5. For example, after \textit{mix coconut milk}, SSM correctly predicts
\textit{add water}, while MAT predicts \textit{put down bowl}. The last two
rows expose complementary limitations: an abrupt \textit{throw packaging}
action is missed by both methods in favour of the temporally plausible
\textit{open microwave}, and SSM predicts the correct verb \textit{stir} but
confuses \textit{potato} with \textit{pan}. These cases show that representative
state relations support longer-range anticipation, while spontaneous events
and fine-grained object identity remain challenging.

Fig.~\ref{fig:compression_tradeoff_main} examines the corresponding
compression--granularity trade-off. Increasing $K$ initially provides richer
temporal cues, with verb and action anticipation peaking at $K=4$; beyond this
point, redundant states weaken temporal discrimination and relational
reasoning. Noun performance instead improves more steadily and saturates from
$K\geq 4$, indicating that object semantics benefit from finer representations
but eventually receive diminishing returns.

The examples in Fig.~\ref{fig:compression_tradeoff_examples_main} reinforce
this asymmetry: the verbs in \textit{stir pan} and \textit{drain vegetables}
are recovered with fewer states, whereas their nouns require $K=6$; for
\textit{place kettle}, finer states recover the noun without consistently
preserving the verb. Thus, compact representations favour action dynamics,
while finer representations retain more object detail but may introduce
redundancy. Joint action anticipation consequently peaks at an intermediate
granularity, motivating adaptive selection that balances temporal abstraction
and fine-grained semantics.

\section{Conclusion}
This work presented SSM, a unified framework for online action detection and anticipation. RSMC compresses the sequence into representative, anchor-conditioned states that serve as compact graph nodes. APL then learns multi-dimensional inter-state relations and performs graph reasoning to model action dynamics and produce a future-oriented graph cue. CTI further models the interaction between past-present cues and the graph cue to refine cross-temporal representations for both tasks. Experiments and ablation studies across multiple benchmarks demonstrate the effectiveness and generalization of the complete framework, particularly the complementary contributions of representative memory compression, relational reasoning, and cross-temporal interaction. Nevertheless, SSM exhibits a compression--granularity trade-off. Overly compact representative states may discard fine-grained object cues and lead to noun-level ambiguity, whereas retaining too many states may introduce redundant information and weaken temporal discrimination in subsequent relational reasoning. Future work will investigate adaptive, semantics-aware state selection to preserve fine-grained object information while maintaining a compact representation of action dynamics.

\section*{Acknowledgments}
This research was funded in part by UKRI EPSRC for the National Edge AI Hub project (EP/Y028813/1).



 


\bibliographystyle{IEEEtran}
\bibliography{tmm_manuscript}

@article{huang2022learning,
  title={Learning representation for clustering via prototype scattering and positive sampling},
  author={Huang, Zhizhong and Chen, Jie and Zhang, Junping and Shan, Hongming},
  journal={IEEE Transactions on Pattern Analysis and Machine Intelligence},
  volume={45},
  number={6},
  pages={7509--7524},
  year={2022},
  publisher={IEEE}
}

@inproceedings{gong2022future,
  title={Future transformer for long-term action anticipation},
  author={Gong, Dayoung and Lee, Joonseok and Kim, Manjin and Ha, Seong Jong and Cho, Minsu},
  booktitle={Proceedings of the IEEE/CVF Conference on Computer Vision and Pattern Recognition},
  pages={3052--3061},
  year={2022}
}

@article{wang2025knowledge,
  title={Knowledge Distillation-Based Spiking Neural Network for Online Video Action Understanding},
  author={Wang, Houlin and Zhang, Shihui and Han, Xueqiang and Pang, Kuo and Zhang, Qixian},
  journal={IEEE Transactions on Industrial Informatics},
  year={2025},
  publisher={IEEE}
}

@inproceedings{yan2026backtrace,
  title={Backtrace Mamba: Reviving Critical Temporal Contexts via Hierarchical Memory Compression for Online Action Detection},
  author={Yan, Su and Li, Jiahua and Wei, Kun and Deng, Cheng},
  booktitle={Proceedings of the AAAI Conference on Artificial Intelligence},
  volume={40},
  number={14},
  pages={11559--11567},
  year={2026}
}

@misc{THUMOS14,
   author = "Jiang, Y.-G. and Liu, J. and Roshan Zamir, A. and Toderici, G. and Laptev,
   I. and Shah, M. and Sukthankar, R.",
   title = "{THUMOS} Challenge: Action Recognition with a Large
   Number of Classes",
   howpublished = "\url{http://crcv.ucf.edu/THUMOS14/}",
   Year = {2014}}

@inproceedings{de2016online,
  title={Online action detection},
  author={De Geest, Roeland and Gavves, Efstratios and Ghodrati, Amir and Li, Zhenyang and Snoek, Cees and Tuytelaars, Tinne},
  booktitle={Computer Vision--ECCV 2016: 14th European Conference, Amsterdam, The Netherlands, October 11-14, 2016, Proceedings, Part V 14},
  pages={269--284},
  year={2016},
  organization={Springer}
}

@inproceedings{wang2021oadtr,
  title={Oadtr: Online action detection with transformers},
  author={Wang, Xiang and Zhang, Shiwei and Qing, Zhiwu and Shao, Yuanjie and Zuo, Zhengrong and Gao, Changxin and Sang, Nong},
  booktitle={Proceedings of the IEEE/CVF International Conference on Computer Vision},
  pages={7565--7575},
  year={2021}
}

@inproceedings{chen2022gatehub,
  title={Gatehub: Gated history unit with background suppression for online action detection},
  author={Chen, Junwen and Mittal, Gaurav and Yu, Ye and Kong, Yu and Chen, Mei},
  booktitle={Proceedings of the IEEE/CVF Conference on Computer Vision and Pattern Recognition},
  pages={19925--19934},
  year={2022}
}

@article{xu2021long,
  title={Long short-term transformer for online action detection},
  author={Xu, Mingze and Xiong, Yuanjun and Chen, Hao and Li, Xinyu and Xia, Wei and Tu, Zhuowen and Soatto, Stefano},
  journal={Advances in Neural Information Processing Systems},
  volume={34},
  pages={1086--1099},
  year={2021}
}

@inproceedings{yang2022colar,
  title={Colar: Effective and efficient online action detection by consulting exemplars},
  author={Yang, Le and Han, Junwei and Zhang, Dingwen},
  booktitle={Proceedings of the IEEE/CVF conference on computer vision and pattern recognition},
  pages={3160--3169},
  year={2022}
}

@article{damen2022rescaling,
  title={Rescaling egocentric vision: Collection, pipeline and challenges for epic-kitchens-100},
  author={Damen, Dima and Doughty, Hazel and Farinella, Giovanni Maria and Furnari, Antonino and Kazakos, Evangelos and Ma, Jian and Moltisanti, Davide and Munro, Jonathan and Perrett, Toby and Price, Will and others},
  journal={International Journal of Computer Vision},
  pages={1--23},
  year={2022},
  publisher={Springer}
}

@article{furnari2020rolling,
  title={Rolling-unrolling lstms for action anticipation from first-person video},
  author={Furnari, Antonino and Farinella, Giovanni Maria},
  journal={IEEE transactions on pattern analysis and machine intelligence},
  volume={43},
  number={11},
  pages={4021--4036},
  year={2020},
  publisher={IEEE}
}

@article{song2018temporal,
  title={Temporal action localization in untrimmed videos using action pattern trees},
  author={Song, Hao and Wu, Xinxiao and Zhu, Bing and Wu, Yuwei and Chen, Mei and Jia, Yunde},
  journal={IEEE transactions on multimedia},
  volume={21},
  number={3},
  pages={717--730},
  year={2018},
  publisher={IEEE}
}

@article{xia2023exploring,
  title={Exploring action centers for temporal action localization},
  author={Xia, Kun and Wang, Le and Shen, Yichao and Zhou, Sanpin and Hua, Gang and Tang, Wei},
  journal={IEEE Transactions on Multimedia},
  volume={25},
  pages={9425--9436},
  year={2023},
  publisher={IEEE}
}

@article{dempster1977maximum,
  title={Maximum likelihood from incomplete data via the EM algorithm},
  author={Dempster, Arthur P and Laird, Nan M and Rubin, Donald B},
  journal={Journal of the royal statistical society: series B (methodological)},
  volume={39},
  number={1},
  pages={1--22},
  year={1977},
  publisher={Wiley Online Library}
}

@article{li2024adaptive,
  title={An adaptive dual selective transformer for temporal action localization},
  author={Li, Qiang and Zu, Guang and Xu, Hui and Kong, Jun and Zhang, Yanni and Wang, Jianzhong},
  journal={IEEE Transactions on Multimedia},
  year={2024},
  publisher={IEEE}
}

@article{li2021restep,
  title={RESTEP into the future: relational spatio-temporal learning for multi-person action forecasting},
  author={Li, Yuke and Wang, Pin and Chan, Ching-Yao},
  journal={IEEE Transactions on Multimedia},
  volume={25},
  pages={1954--1963},
  year={2021},
  publisher={IEEE}
}

@article{zhou2024smc,
  title={SMC-NCA: Semantic-guided Multi-level Contrast for Semi-supervised Temporal Action Segmentation},
  author={Zhou, Feixiang and Jiang, Zheheng and Zhou, Huiyu and Li, Xuelong},
  journal={IEEE Transactions on Multimedia},
  year={2024},
  publisher={IEEE}
}

@article{liu2023hcm,
  title={Hcm: Online action detection with hard video clip mining},
  author={Liu, Siyu and Cheng, Jian and Xia, Ziying and Xi, Zhilong and Hou, Qin and Dong, Zhicheng},
  journal={IEEE Transactions on Multimedia},
  volume={26},
  pages={3626--3639},
  year={2023},
  publisher={IEEE}
}

@inproceedings{wang2023memory,
  title={Memory-and-anticipation transformer for online action understanding},
  author={Wang, Jiahao and Chen, Guo and Huang, Yifei and Wang, Limin and Lu, Tong},
  booktitle={Proceedings of the IEEE/CVF International Conference on Computer Vision},
  pages={13824--13835},
  year={2023}
}

@inproceedings{furnari2019would,
  title={What would you expect? anticipating egocentric actions with rolling-unrolling lstms and modality attention},
  author={Furnari, Antonino and Farinella, Giovanni Maria},
  booktitle={Proceedings of the IEEE/CVF International conference on computer vision},
  pages={6252--6261},
  year={2019}
}

@inproceedings{xu2019temporal,
  title={Temporal recurrent networks for online action detection},
  author={Xu, Mingze and Gao, Mingfei and Chen, Yi-Ting and Davis, Larry S and Crandall, David J},
  booktitle={Proceedings of the IEEE/CVF international conference on computer vision},
  pages={5532--5541},
  year={2019}
}

@article{vaswani2017attention,
  title={Attention is all you need},
  author={Vaswani, Ashish and Shazeer, Noam and Parmar, Niki and Uszkoreit, Jakob and Jones, Llion and Gomez, Aidan N and Kaiser, {\L}ukasz and Polosukhin, Illia},
  journal={Advances in neural information processing systems},
  volume={30},
  year={2017}
}

@article{zhao2022progressive,
  title={Progressive privileged knowledge distillation for online action detection},
  author={Zhao, Peisen and Xie, Lingxi and Wang, Jiajie and Zhang, Ya and Tian, Qi},
  journal={Pattern Recognition},
  volume={129},
  pages={108741},
  year={2022},
  publisher={Elsevier}
}

@article{cho2014learning,
  title={Learning phrase representations using RNN encoder-decoder for statistical machine translation},
  author={Cho, Kyunghyun and Van Merri{\"e}nboer, Bart and Gulcehre, Caglar and Bahdanau, Dzmitry and Bougares, Fethi and Schwenk, Holger and Bengio, Yoshua},
  journal={arXiv preprint arXiv:1406.1078},
  year={2014}
}

@inproceedings{eun2020learning,
  title={Learning to discriminate information for online action detection},
  author={Eun, Hyunjun and Moon, Jinyoung and Park, Jongyoul and Jung, Chanho and Kim, Changick},
  booktitle={Proceedings of the IEEE/CVF conference on computer vision and pattern recognition},
  pages={809--818},
  year={2020}
}

@inproceedings{kitani2012activity,
  title={Activity forecasting},
  author={Kitani, Kris M and Ziebart, Brian D and Bagnell, James Andrew and Hebert, Martial},
  booktitle={Computer Vision--ECCV 2012: 12th European Conference on Computer Vision, Florence, Italy, October 7-13, 2012, Proceedings, Part IV 12},
  pages={201--214},
  year={2012},
  organization={Springer}
}

@inproceedings{deo2017learning,
  title={Learning and predicting on-road pedestrian behavior around vehicles},
  author={Deo, Nachiket and Trivedi, Mohan M},
  booktitle={2017 IEEE 20th International Conference on Intelligent Transportation Systems (ITSC)},
  pages={1--6},
  year={2017},
  organization={IEEE}
}

@inproceedings{yang2024online,
  title={Online Mouse Behavior Detection by Historical Dependency and Typical Instances},
  author={Yang, Xinyu and Zhou, Feixiang and Zhou, Huiyu},
  booktitle={ICASSP 2024-2024 IEEE International Conference on Acoustics, Speech and Signal Processing (ICASSP)},
  pages={3990--3994},
  year={2024},
  organization={IEEE}
}

@article{zhou2025cross,
  title={Cross-skeleton interaction graph aggregation network for representation learning of mouse social behaviour},
  author={Zhou, Feixiang and Yang, Xinyu and Chen, Fang and Chen, Long and Jiang, Zheheng and Zhu, Hui and Heckel, Reiko and Wang, Haikuan and Fei, Minrui and Zhou, Huiyu},
  journal={IEEE Transactions on Image Processing},
  year={2025},
  publisher={IEEE}
}

@inproceedings{osman2021slowfast,
  title={Slowfast rolling-unrolling lstms for action anticipation in egocentric videos},
  author={Osman, Nada and Camporese, Guglielmo and Coscia, Pasquale and Ballan, Lamberto},
  booktitle={Proceedings of the IEEE/CVF International Conference on Computer Vision},
  pages={3437--3445},
  year={2021}
}

@inproceedings{roy2024interaction,
  title={Interaction region visual transformer for egocentric action anticipation},
  author={Roy, Debaditya and Rajendiran, Ramanathan and Fernando, Basura},
  booktitle={Proceedings of the IEEE/CVF Winter Conference on Applications of Computer Vision},
  pages={6740--6750},
  year={2024}
}

@inproceedings{liu2022hybrid,
  title={A hybrid egocentric activity anticipation framework via memory-augmented recurrent and one-shot representation forecasting},
  author={Liu, Tianshan and Lam, Kin-Man},
  booktitle={Proceedings of the IEEE/CVF Conference on Computer Vision and Pattern Recognition},
  pages={13904--13913},
  year={2022}
}

@article{qi2021self,
  title={Self-regulated learning for egocentric video activity anticipation},
  author={Qi, Zhaobo and Wang, Shuhui and Su, Chi and Su, Li and Huang, Qingming and Tian, Qi},
  journal={IEEE transactions on pattern analysis and machine intelligence},
  volume={45},
  number={6},
  pages={6715--6730},
  year={2021},
  publisher={IEEE}
}

@inproceedings{zhao2022real,
  title={Real-time online video detection with temporal smoothing transformers},
  author={Zhao, Yue and Kr{\"a}henb{\"u}hl, Philipp},
  booktitle={European Conference on Computer Vision},
  pages={485--502},
  year={2022},
  organization={Springer}
}

@inproceedings{pang2025context,
  title={Context-enhanced memory-refined transformer for online action detection},
  author={Pang, Zhanzhong and Sener, Fadime and Yao, Angela},
  booktitle={Proceedings of the Computer Vision and Pattern Recognition Conference},
  pages={8700--8710},
  year={2025}
}

@article{luo2022learning,
  title={Learning multi-dimensional edge feature-based au relation graph for facial action unit recognition},
  author={Luo, Cheng and Song, Siyang and Xie, Weicheng and Shen, Linlin and Gunes, Hatice},
  journal={arXiv preprint arXiv:2205.01782},
  year={2022}
}

@article{bresson2017residual,
  title={Residual gated graph convnets},
  author={Bresson, Xavier and Laurent, Thomas},
  journal={arXiv preprint arXiv:1711.07553},
  year={2017}
}

@inproceedings{foo2024action,
  title={Action Detection via an Image Diffusion Process},
  author={Foo, Lin Geng and Li, Tianjiao and Rahmani, Hossein and Liu, Jun},
  booktitle={Proceedings of the IEEE/CVF Conference on Computer Vision and Pattern Recognition},
  pages={18351--18361},
  year={2024}
}

@inproceedings{xie2024towards,
  title={Towards understanding future: Consistency guided probabilistic modeling for action anticipation},
  author={Xie, Zhao and Shi, Yadong and Wu, Kewei and Cheng, Yaru and Guo, Dan},
  booktitle={Proceedings of the AAAI Conference on Artificial Intelligence},
  volume={38},
  number={6},
  pages={6243--6251},
  year={2024}
}

@inproceedings{girdhar2021anticipative,
  title={Anticipative video transformer},
  author={Girdhar, Rohit and Grauman, Kristen},
  booktitle={Proceedings of the IEEE/CVF international conference on computer vision},
  pages={13505--13515},
  year={2021}
}

@article{wang2024contextdet,
  title={ContextDet: Temporal Action Detection with Adaptive Context Aggregation},
  author={Wang, Ning and Xiao, Yun and Peng, Xiaopeng and Chang, Xiaojun and Wang, Xuanhong and Fang, Dingyi},
  journal={arXiv preprint arXiv:2410.15279},
  year={2024}
}

@inproceedings{diko2024semantically,
  title={Semantically guided representation learning for action anticipation},
  author={Diko, Anxhelo and Avola, Danilo and Prenkaj, Bardh and Fontana, Federico and Cinque, Luigi},
  booktitle={European Conference on Computer Vision},
  pages={448--466},
  year={2024},
  organization={Springer}
}

@inproceedings{wu2022memvit,
  title={Memvit: Memory-augmented multiscale vision transformer for efficient long-term video recognition},
  author={Wu, Chao-Yuan and Li, Yanghao and Mangalam, Karttikeya and Fan, Haoqi and Xiong, Bo and Malik, Jitendra and Feichtenhofer, Christoph},
  booktitle={Proceedings of the IEEE/CVF Conference on Computer Vision and Pattern Recognition},
  pages={13587--13597},
  year={2022}
}

@inproceedings{guo2024uncertainty,
  title={Uncertainty-aware Action Decoupling Transformer for Action Anticipation},
  author={Guo, Hongji and Agarwal, Nakul and Lo, Shao-Yuan and Lee, Kwonjoon and Ji, Qiang},
  booktitle={Proceedings of the IEEE/CVF Conference on Computer Vision and Pattern Recognition},
  pages={18644--18654},
  year={2024}
}

\vfill

\end{document}